\documentclass[runningheads,oribibl]{llncs}

\usepackage{url}
\usepackage{graphicx}
\usepackage[cp1250]{inputenc}
\usepackage{url}
\usepackage[ruled,vlined]{algorithm2e}
\usepackage{array}
\usepackage{amsmath}
\newcolumntype{x}[1]{>{\raggedleft\hspace{0pt}}p{#1}}

\newenvironment{keywords}{
       \list{}{\advance\topsep by0.35cm\relax\small
       \leftmargin=1cm
       \labelwidth=0.35cm
       \listparindent=0.35cm
       \itemindent\listparindent
       \rightmargin\leftmargin}\item[\hskip\labelsep
                                     \bfseries Keywords:]}
     {\endlist}

\begin{document}

\bibliographystyle{ieeetr}

\title{Random Graphs for Performance Evaluation of Recommender Systems}
\author{Szymon Chojnacki and Mieczysław Kłopotek}

\institute{Institute of Computer Science, \\Polish Academy of Sciences \\ J.K. Ordona 21, 01-237 Warsaw, Poland\\
 \email{\{sch, klopotek\}@ipipan.waw.pl}}

\maketitle

\begin{abstract}
The purpose of this article is to introduce a new analytical framework dedicated to measuring performance of recommender systems. The standard approach is to assess the quality of a system by means of accuracy related statistics. However, the specificity of the environments in which recommender systems are deployed requires to pay much attention to speed and memory requirements of the algorithms. Unfortunately, it is implausible to assess accurately the complexity of various algorithms with formal tools. This can be attributed to the fact that such analyses are usually based on an assumption of dense representation of underlying data structures. Whereas, in real life the algorithms operate on sparse data and are implemented with collections dedicated for them. Therefore, we propose to measure the complexity of recommender systems with artificial datasets that posses real-life properties. We utilize recently developed bipartite graph generator to evaluate how state-of-the-art recommender systems' behavior is determined and diversified by topological properties of the generated datasets.

\end{abstract}

\section{Introduction}

Recommender systems are an important component of the Intelligent Web. The systems make information retrieval easier and push users from typing queries towards clicking at suggested links. We experience real-life recommender systems when browsing for books, movies, news or music. The engines are an essential part of such websites as \textit{Amazon}, \textit{MovieLens} or \textit{Last.fm}. Recommender systems are used to deal with the tasks that are typical for statistical classification methods. They fit especially the scenarios in which the number of attributes, classes or missing values is large. Classic data-mining techniques like \textit{logistic regression} or \textit{decision trees} are well suited to predict which category of news is the most interesting for a particular customer. Recommender systems are used to output more fine-grained results and point at concrete stories. 

In recent years we have observed a surge of interest of research community in recommender systems. One of the events that was responsible for this phenomenon was the Netflix Prize challenge \cite{netflix}. The competition was organized by a large DVD retailer in US. The prize of 1 million dollars was awarded to the team that managed to improve RMSE (root mean standard error) of the retailer's Cinematch algorithm by more than 10\%. It turned out that classic collaborative filtering techniques \cite{herlocker} do not perform as good as SVD-based (Singular Value Decomposition) approaches \cite{koren}. During the competiotion a new method derived from the field of artificial neural networks\footnote{Recently new algorithm \cite{duch} based on the concept of \textit{k-separability} was developed with promising results, which shows that the power of artificial neural networks in this domain has not been fully harnessed yet.} was applied with high accuracy (i.e. Restricted Boltzmann Machines \cite{rbm}). The most successful solution was achieved by blending various algorithms \cite{koren2}. 

The lesson we learned during the Netflix Prize is that the difference between the quality of simple methods and the sophisticated ones is not as significant as we could have expected. Moreover, in order to lower RMSE an ensemble of complex and computationally intensive methods has to be done. Even though the organizers made much effort to deliver realistic and huge data, the setting did not envision the problems that we need to face in diverse real-life recommender systems applications, such as:
\begin{itemize}
	\item the \textit{Cold Start} problem, i.e. an arrival of new users with short history (e.g. restricted to the last HTTP session)
	\item instant creation of new items (e.g. news, auction items or photos)
	\item real-time feedback from users about our performance
\end{itemize}

These drawbacks were overcome during the Online Task of the Discovery Challenge \cite{jaeschke2009} organized as a part of the ECML 2009 (European Conference on Machine Learning). The owners of the \url{www.BibSonomy.org} bookmarking portal opened its interfaces to recommender systems taking part in the evaluation. Whenever a user of BibSonomy was bookmarking a digital resource (a publication or a website) a query was sent to all the systems. The tag recommendation of a random one was displayed to the user. After the action a feedback with user's actions was sent to all systems. The systems could have been maintained during the challenge, because they were configured as web services. The results showed that all of the teams found it difficult to deliver majority of its recommendations within time constraint of $1~000$ ms. 

Our research was motivated by the above result and an observation that the development of recommender systems is limited by a fact that there are not enough possibilities to test the algorithms with various datasets. The data structure used by recommender systems is a sparse $user \times item$ matrix with ratings. It is a hard exercise to generate randomly such matrices with predefined properties resembling real-life situations, because of three reasons. Firstly, if we fix the number of users, items and rankings and try to place the rankings randomly in the matrix we obtain symmetric distribution of the number of items rated by users (and vice versa). However, in real-life datasets the distributions are skewed. Secondly, simple random selection results in no correlations among user's preferences. Such correlations exist in real datasets. Thirdly, if we generate one matrix with some desired properties and would like to add new users or items, we would probably loose the properties of the original matrix.
 
We have challenged the problem recently \cite{chojnacki_CC}. We proposed to look at the matrix with ratings as if it was a bipartite graph with nodes of both modalities representing users and items respectively. A rating from the matrix is mapped onto an edge in the bigraph. We proposed an algorithm in which we can control not only simple statistics like numbers of users, items or rankings, but also obtain skewed distributions and correlations among users or items. Moreover, our random bigraph generator's asymptotic properties were verified by virtue of formal and numerical tools and we can add user or items to the graph without loosing the properties of the original datasets. The algorithm was obtained by adapting the advances in unipartite complex networks modeling onto a bipartite ground. We modified the preferential attachment model of Barab\'{a}si and Albert \cite{bar99a} with the extension of Liu \cite{liu_2002} and generalized the \textit{surfing mechanizm} by V\'azquez \cite{vazquez}.

In this paper we apply the generator to produce several random bigraphs with various properties and evaluate how the properties impinge on the performance of analyzed recommender systems. We analyze four features of the systems that in our opinion are responsible for the success of an algorithm in a real-life setting:
\begin{itemize}
	\item time required to build a model from a scratch
	\item memory consumption of the trained model
	\item latency of creating a recommendation
	\item time of updating the model with new ratings
\end{itemize}
We considered six algorithms during the tests: UserBased, ItemBased, SlopeOne, UserThreshold, KnnItem and SVD \cite{mahoutinaction}. We relied on high-performance Mahout library \cite{mahout}. Our attention was focused on five properties of artificial datasets: (1) size of the graph, (2) relative number of edges, (3) proportion of the number of users to the number of items, (4) clustering of edges, (5) distributions of node degrees of both modalities. We also checked the influence on the performance UserBased model of two characteristics: similarity measure and the size of the neighborhood.
 
Throughout the article we show that our approach can be used to better understand the features of datasets that are responsible for the performance of recommender systems. We utilize this knowledge to show (1) which algorithms are best suited for various scenarios, (2) identify datasets' features that are correlated with improving performance of some algorithms and diminishing performance of others, (3) point at potential directions of improving the implementations of the algorithms.  

The rest of the article is organized as follows. In Section 2 we describe how recommender systems are evaluated. In Section 3 we outline the details of applied random bigraph generator. The fourth section contains the results of extensive experiments. The last fifth section is dedicated for the concluding remarks. 

%\begin{figure}[htbp]
%	\centering
%		\includegraphics[width=\textwidth]{motivation.pdf}
%\end{figure}

\section{Performance of Recommender Systems}

\begin{figure}[htbp]
%\begin{tabular}{cc}
\centering
	\includegraphics[width=0.9\textwidth]{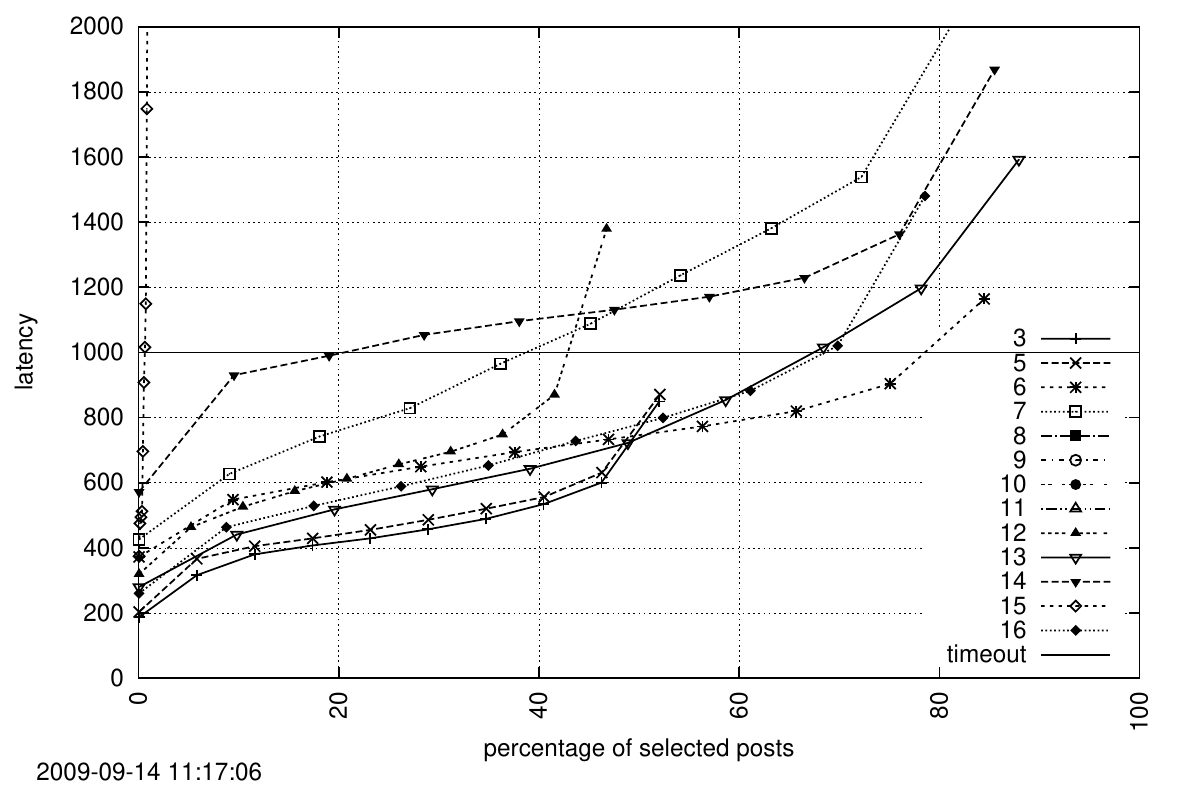} %&\includegraphics[width=0.5\textwidth]{ecml_L11.pdf}\\
%	\end{tabular}
	\caption{Latency of recommender systems evaluated during the Online Task of ECML-PKDD 2009 Discovery Challenge. None of the systems delivered recommendations to all queries. The most recommendations were output by system number 13 (over 80\%). The points on curves show the percentiles of latency for each algorithm. For example the second most right point point of algorithm 13 indicates that almost 80\% of its recommendations were delivered within $1~200$ms. Source: http://www.kde.cs.uni-kassel.de/ws/dc09/results/online/.}
	\label{fig:ecml}
\end{figure}

In our research we perceive performance in terms of real-life speed of using an algorithm. We omit analysis of statistical indicators such as accuracy, recall, f-measure, RMSE or lift charts. This is because, even though they are crucial in the process of selecting an algorithm to be deployed, it seems pointless to evaluate these measures within randomly generated datasets. One could argue that the usefulness of artificial datasets is questionable. And the best strategy is to evaluate all possible algorithms with one's real dataset and choose the most accurate that gives recommendations within specified time constraints. We argue with this point of view. Based on our experience we are more likely to believe that the structure and topology of datasets changes rapidly and the performance may be affected by appearance of outlying observations or unexpected growth in scale. It is hard to foresee all potential pitfalls and the need to evaluate the algorithms within a wide range of artificial data emerges. In order to justify our deduction we present in Fig. \ref{fig:ecml} the results of an online evaluation of the systems that participated in the social bookmarking Discovery Challenge organized as a part of ECML'09.

The algorithms were deployed as web services and latency was measured as the difference of time between sending a request and receiving a list of top five recommended tags. We participated in the evaluation and managed to lower the latency of our system to the level of $400$ ms \cite{hsi}. It turned out to be an important improvement for the users of the BibSonomy and resulted in the highest rate of clicks among all the evaluated systems. 

A good starting point when analyzing complexity of any algorithm is to assess its asymptotic properties. Such analyses are usually based on several simplifying assumptions. One of the assumptions that is virtually never met is the dense data structure representation assumption. For example vectors are either implemented as dense (e.g. \verb!double[]!) or sparse arrays (e.g. \verb!ArrayList<Double>!). Dedicated collections are used to find optimal trade-offs between memory consumption and speed. Moreover, several advanced issues such as Java \verb!Object! overhead and its influence on garbage collectors performance need to be taken into account. A great discussion about these problems is contained in chapter 3 of \cite{mahoutinaction}. 

 Nonetheless, theoretical analysis of complexity strengthens our general intuition about the upper bound of time needed for computations. A thorough analysis of wide range of recommender algorithms is described in \cite{toscher}. In particular, it is assessed that training time, latency and memory consumption of SVD-based recommender are $O(E)$, $O(1)$, $O(U+I)$, where $E$ is the number of ratings, $U$ is the number of users and $I$ is the number of items.  The limitation of this kind of analysis is the fact that except of some rare situations it does not help us to decide which of two selected algorithms is expected to perform better on real data. An example of a set of soft rules trying to cope with such questions is drawn in Table \ref{tabka}.

\begin{table}[h]
	\centering
		\begin{tabular}{p{5cm}cp{7cm}}
		\hline
		Recommender & & Key features\\
		\hline
		UserBased \cite{herlocker} & & Fast when number of users is relatively small \\
		
		ItemBased \cite{item} & & Fast when number of items is relatively small\\
		
		SlopeOne \cite{slopeone} & & Recommendations and updates fast at runtime\\
							& & Requires large precomputations\\
							& & Suitable when number of items is relatively small\\
		KnnItem \cite{knn} & & Good when number of items is relatively smaller\\
		
		SVD \cite{svd} & & Good results\\
							& & Requires large precomputations\\
			\hline
				&&\\
		\end{tabular}
	\caption{Comparison of recommender systems,  based on \cite{mahoutinaction}.}
	\label{tabka}
\end{table}
 
In our experiments we did not find evidence to assert that UserBased algorithm performs significantly different than ItemBased when the proportion of the number of users to the number of items varies. We also found that there exist other factors than $U$, $I$ or $E$ that are interpretable and impinge on the performance in a coherent way. The results are discussed in detail in Sec. 4.

\newpage
\section{Bipartite Random Graph Generator}

In this section we describe an algorithm used to generate random bigraphs. The algorithm was introduced and described in detail in \cite{chojnacki_CC}. In Sec. 3.1 we define the parameters of the algorithm, in Sec. 3.2 the properties of the generator are outlined. 

\subsection{Generative procedure}

The generative procedure consists of three steps: (1) new node creation, (2) edge attachment type selection and (3) running bouncing mechanism. The steps are run after an initialization of the bigraph. The procedure requires specifying eight parameters:
\newline
\begin{tabular}{lcl}
$m$ & - & the number of initial loose edges with a user and an item at the ends\\
$T$ & - & the number of iterations\\
$p$ & - & the probability that a new node is a user\\
    &  & $(1-p)$ is the probability that a new node is an item\\
$u$  & - & the number of edges created by each new user\\
$v$  & - & the number of edges created by each new item\\
$\alpha$  & - & the probability that a new user's edge is being connected to\\
      &  & an item with preferential attachment\\
$\beta$  & - & the probability that a new item's edge is being connected to\\
      &  & a user with preferential attachment\\
$b$ & - & the fraction of preferentially attached edges \\
&&that were created via a \textit{bouncing mechanism} \\

\end{tabular}

In the preferential attachment mechanism the probability that a node is drawn is linearly proportional to its degree. Opposite to the preferential attachment is random attachment, in which a probability of selection is equal for all nodes. The model is based on an iterative repetition of three steps.
\newline
\textbf{Step 1} If a random number is greater then $p$ create a new user with $u$ loose edges, otherwise create a new item with $v$ loose edges.
\newline
\textbf{Step 2} For each edge decide whether to join it to a node of the second modality randomly or with preferential attachment. The probability of selection preferential attachment is $\alpha$ for new user and $\beta$ for new item.
\newline
\textbf{Step 3} For each edge that is supposed to be created with preferential attachment decide if it should also be generated via a bouncing mechanism. 

\begin{figure}[htbp]
	\centering
		\includegraphics[width=\textwidth]{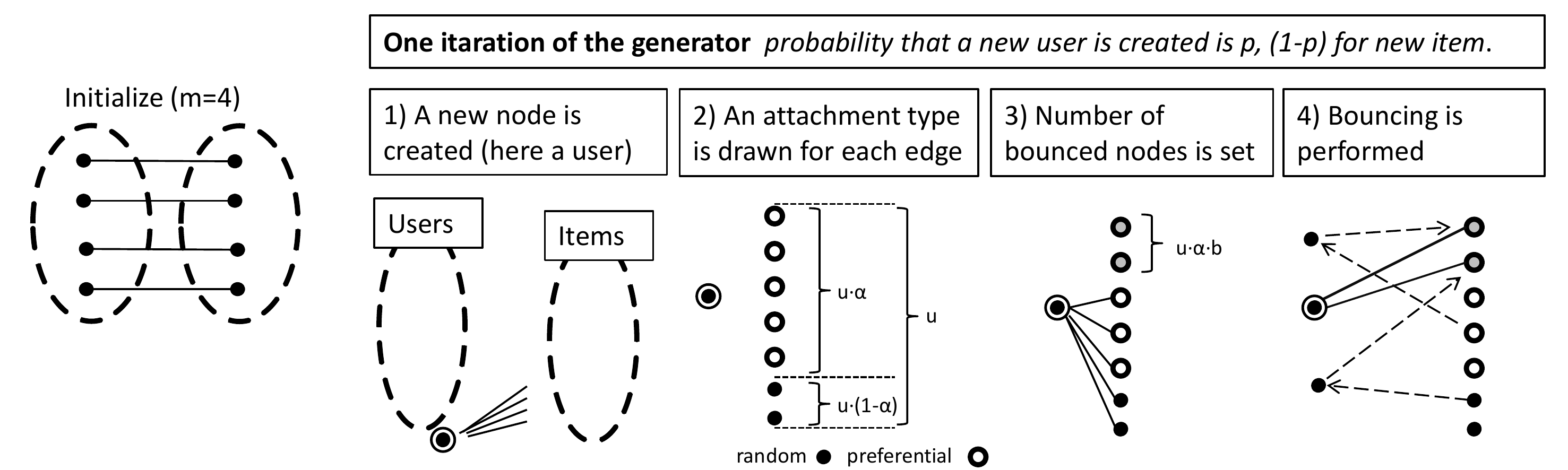}
	\caption{For each edge of a new node, that is to be connected with an existing node with accordance to the preferential attachment mechanism, a decision is made whether to create it via a bouncing mechanism. In case of attaching new user node, $u$ new edges are created. On average $u\cdot\alpha$ edges' endings are to be drawn preferentially and $u\cdot\alpha\cdot b$ of them are to be obtained via bouncing from the nodes that are already selected. }
	\label{fig:steps}
\end{figure}

Bouncing is performed in three micro steps: (1) a random node is drawn from the nodes that are already joined with the new node, (2) a random neighbor of the drawn node is chosen, (3) a random neighbor of the neighbor is selected for joining with the new node. The bouncing mechanism was injected into the model in order to parametrize the level of transitivity in a graph. The transitivity is a feature of real datasets and in terms of recommender systems represent the correlations between items ranked by different users. In unipartite graphs transitivity is measured by the \textit{local clustering coefficient}, which is calculated for each node as a number of edges among direct neighbors of the node divided by all possible pairs of the neighbors. In bipartite graphs the coefficient is always zero and is substituted by \textit{bipartite local clustering coefficient (BLCC)} \cite{chojnacki_CC}. Bipartite local clustering coefficient  of node $j$ takes values of one minus the proportion of node's second neighbors to the potential number of the second neighbors of the node. The value of $BLCC$ calculated for node $j$ is given by: 

\begin{equation}
BLCC_j=1-\frac{ |N_2(j)|}
{\sum_{i \in N_1(j)}{(k_i - 1)}},
\label{blcc}
\end{equation}

\noindent where $|N_2(j)|$ stands for the number of the second neighbors of node $j$, $N_1(j)$ is a set of the first neighbors of node $j$ and $k_i$ is a degree of node $i$. The steps of the generator are depicted in Fig. \ref{fig:steps}

\subsection{Properties}

One can see that after $t$ iterations the bigraph consists of $|U(t)|=m+p  t$ users, $|I(t)|=m+(1-p)  t$ items, and $|E(t)|=m+t  (p  u+(1-p)  v)$ edges. Let's denote by $\eta$ an average number of edges created during one iteration $\eta=(p  u+(1-p)  v)$. After relatively many iterations $(t>>m)$ we can neglect $m$. In the presented model, an average user degree is:

$$\frac{|E(t)|}{|U(t)|}=\frac{m+t  (p  u+(1-p)  v)}{m+p  t}\approx \frac{\eta}{p},$$

\noindent Analogously an average item degrees is:

$$\frac{|E(t)|}{|I(t)|} \approx  \frac{\eta}{(1-p)}.$$

\begin{figure}[htbp]
\begin{tabular}{lr}
	\centering
\includegraphics[width=0.5\textwidth]{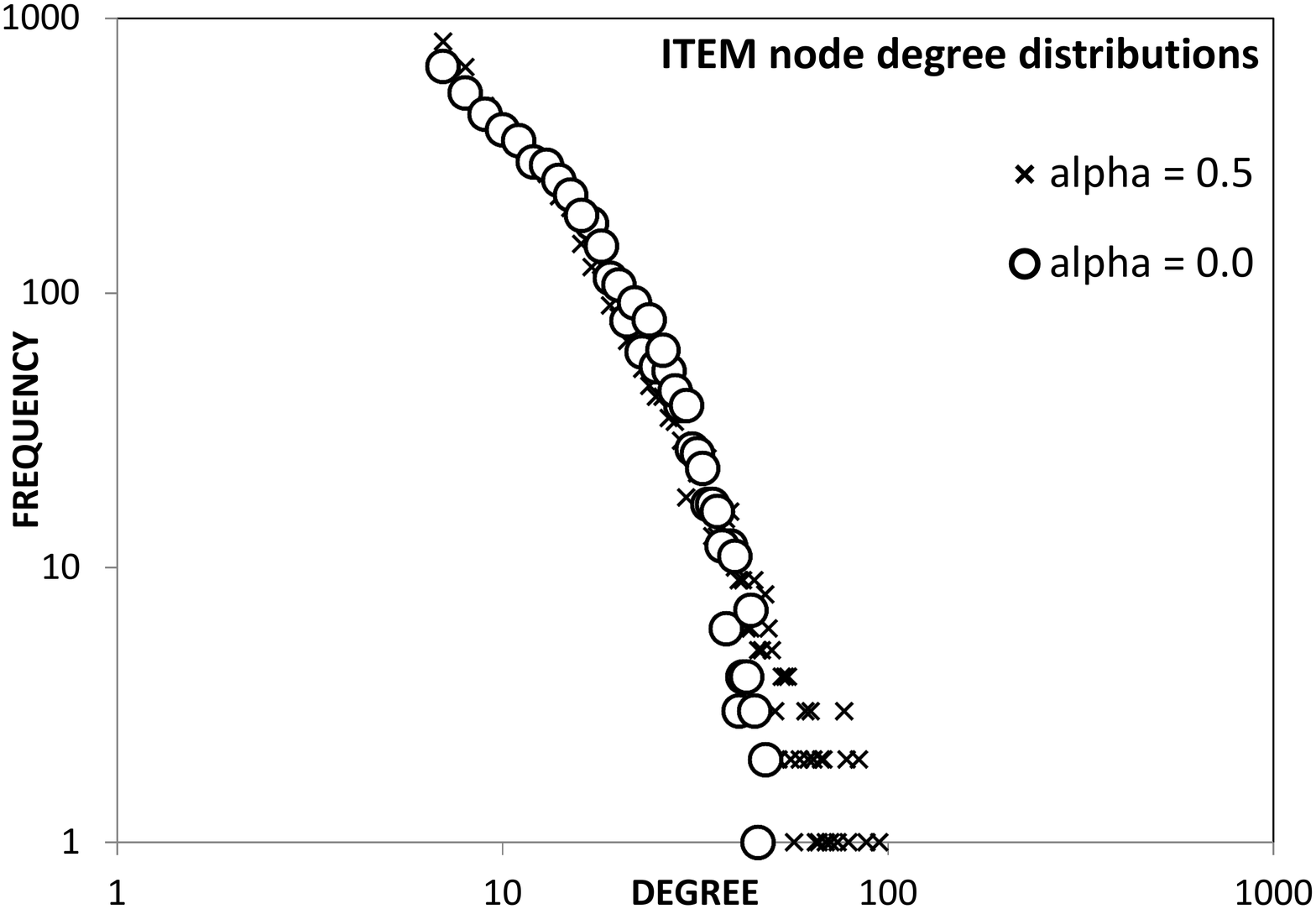}&\includegraphics[width=0.5\textwidth]{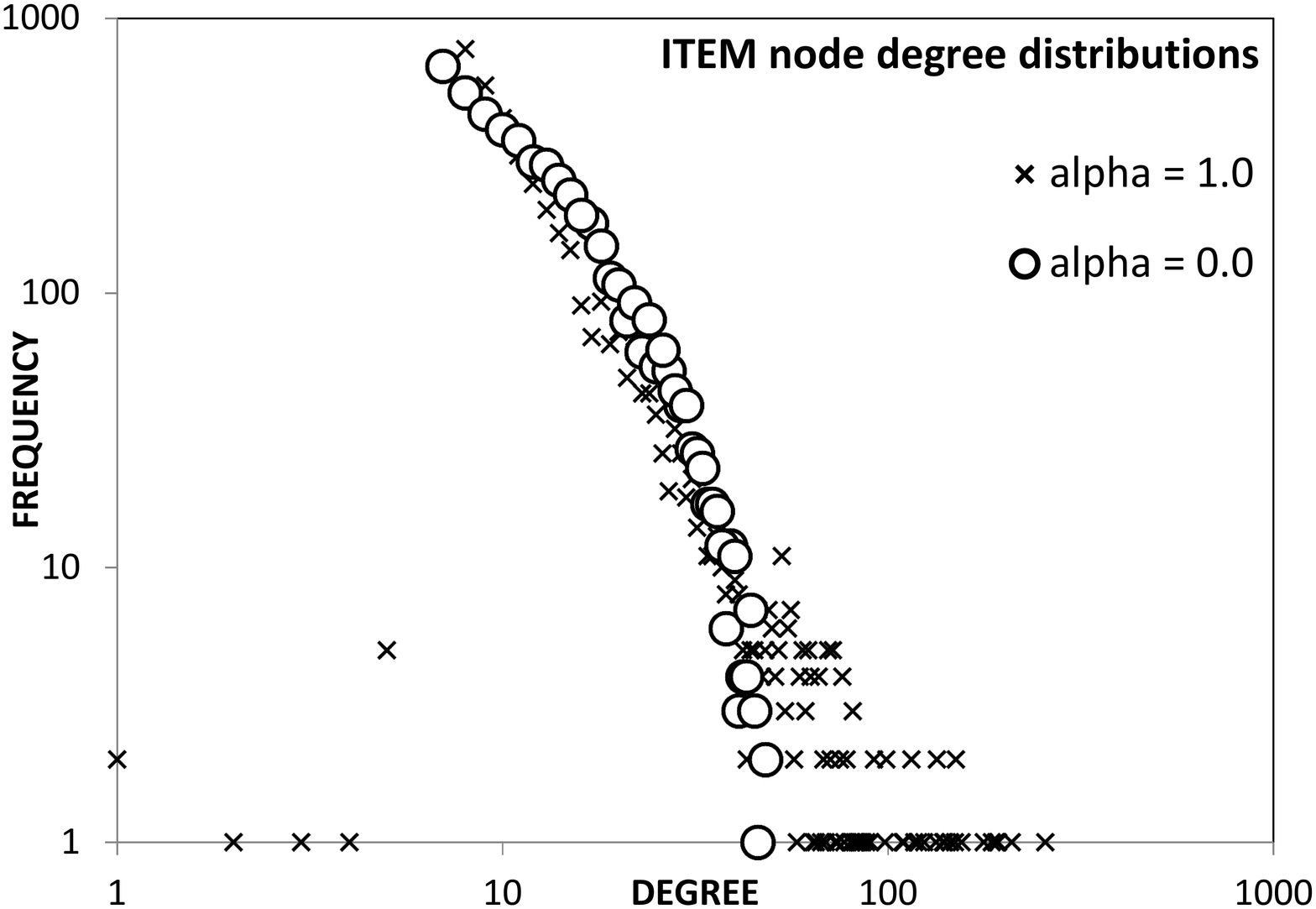}\\
	\end{tabular}
	\caption{ Parameters $\alpha$ and $\beta$ enable us to control the shape of node degree distributions. As the values approach unity, the shape becomes power-law, as the values vanish to zero the shape tends to exponential. }

	\label{fig:distrUSER_ITEM}
\end{figure}

\begin{figure}[htbp]
\begin{tabular}{cc}
	%\centering
	\includegraphics[width=0.5\textwidth]{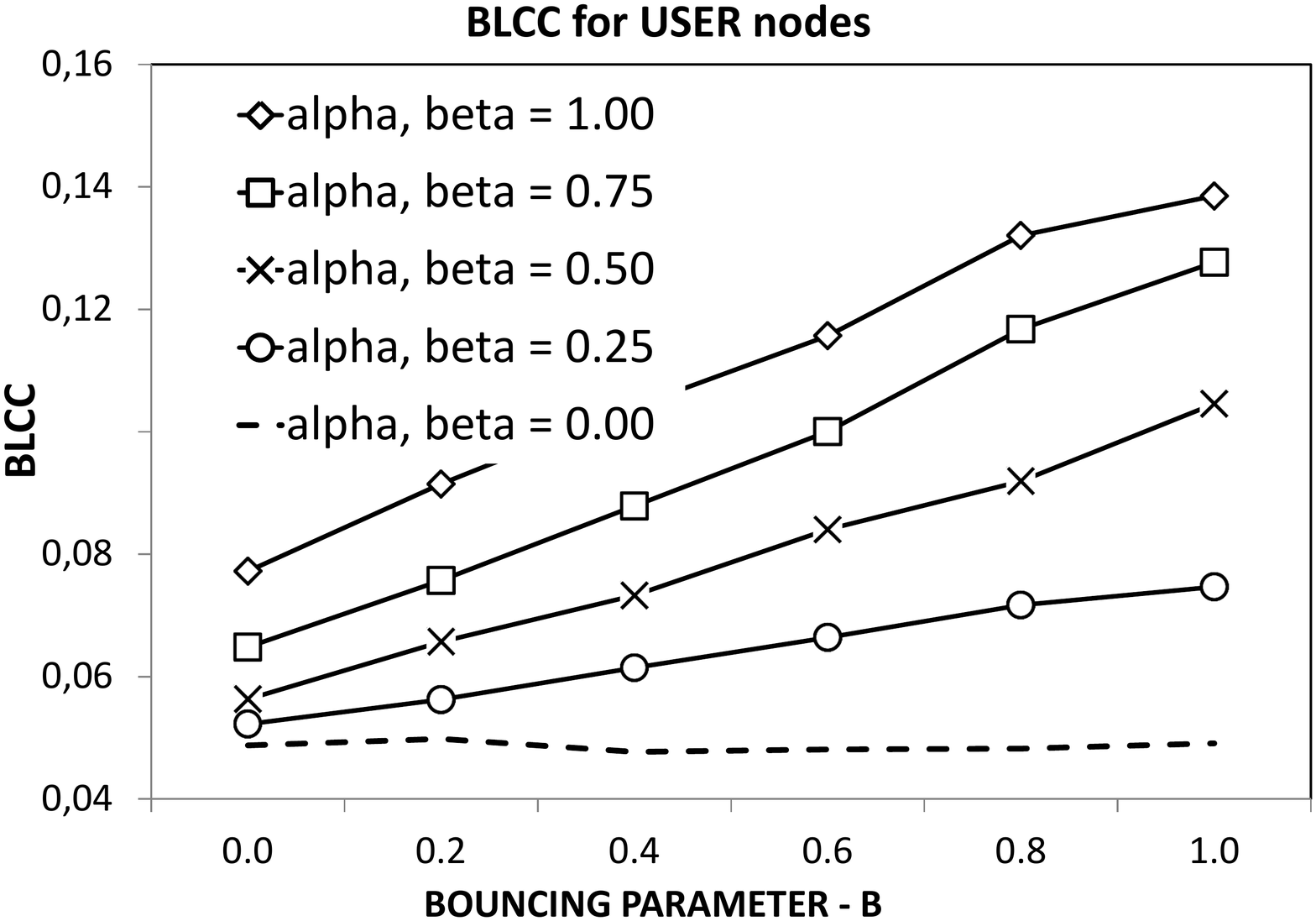}&\includegraphics[width=0.5\textwidth]{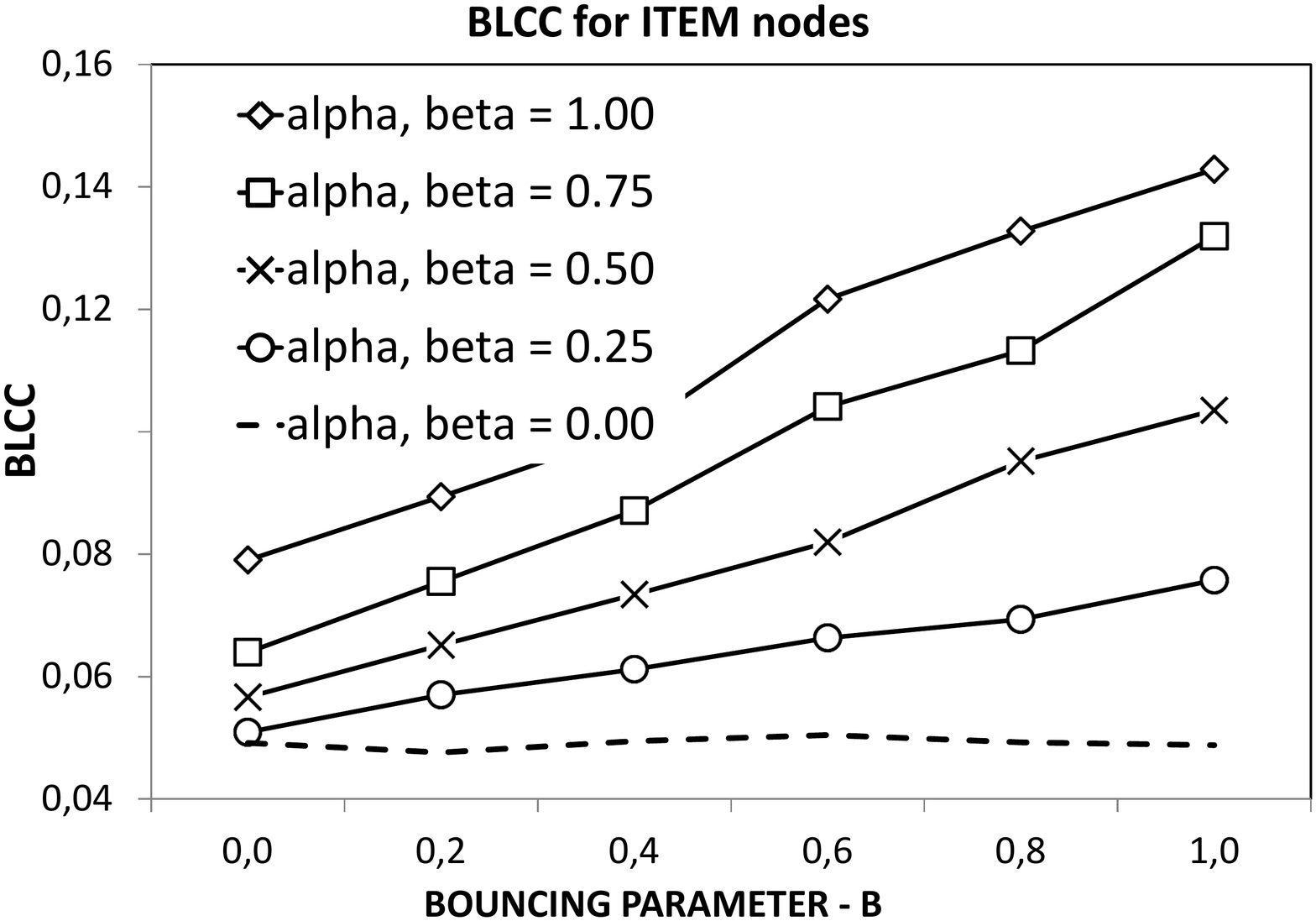}\\
	\end{tabular}
	\caption{The growth of parameter \textit{b} results in higher values of BLCC. It does not influence BLCC if $\alpha=\beta=0.0$.}
	\label{fig:BLCC1}
\end{figure}

\begin{figure}[htbp]
\begin{tabular}{cc}
	%\centering
	\includegraphics[width=0.5\textwidth]{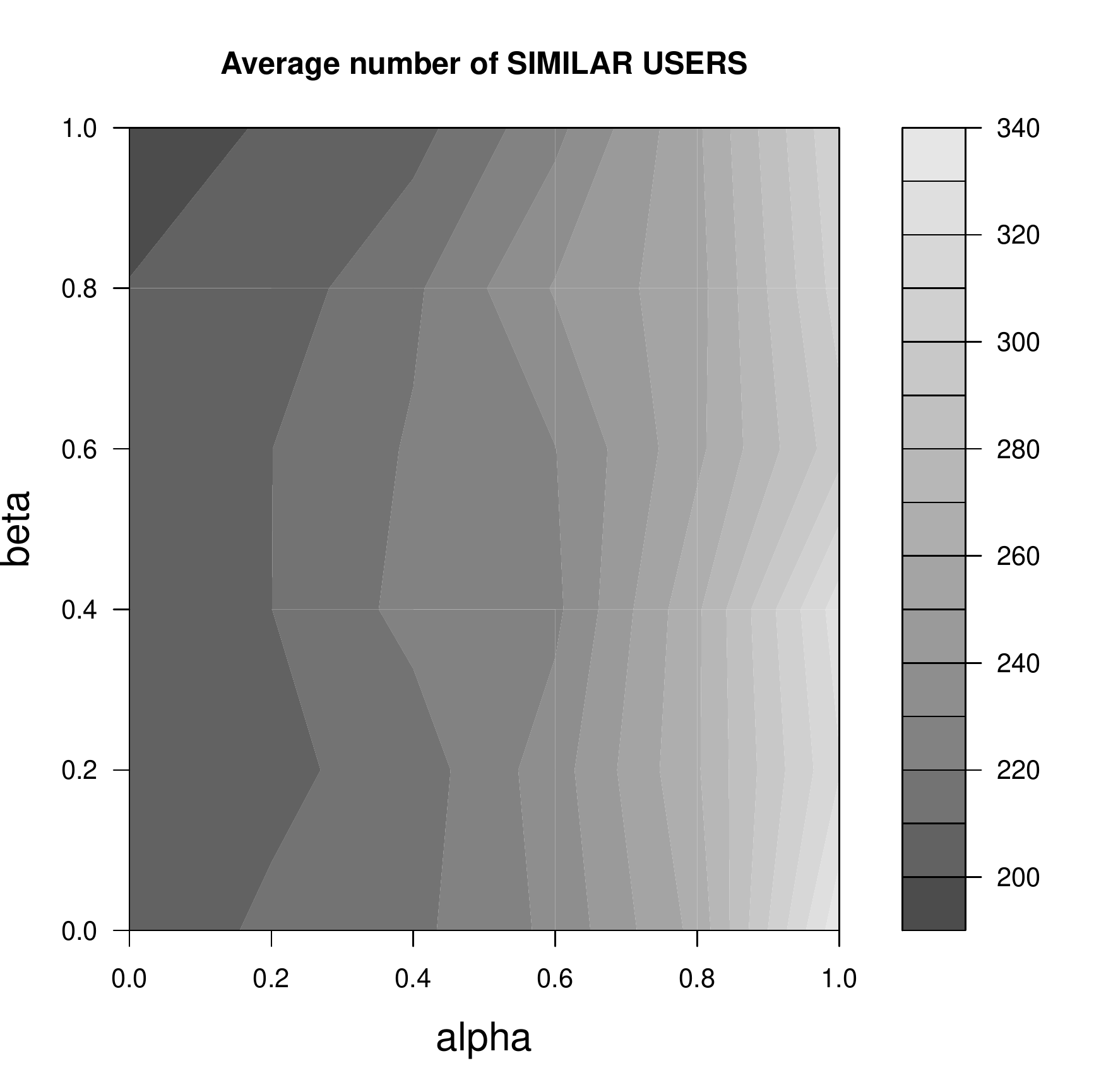}&\includegraphics[width=0.5\textwidth]{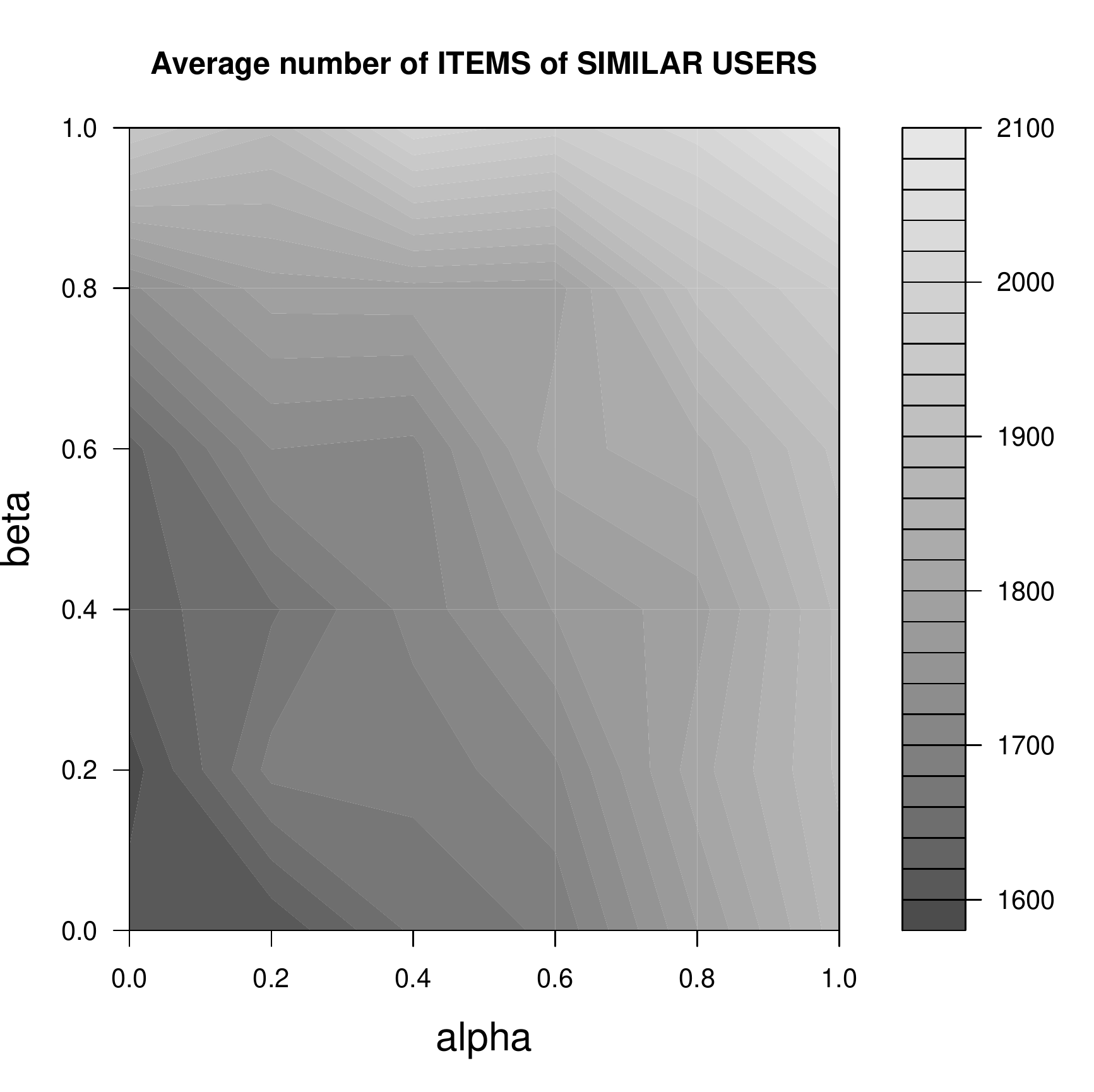}\\
	\end{tabular}	\caption{ The average number of similar users (having ranked at least one item in common with a defined user) and their items depends on both $\alpha$ and $\beta$.}
	\label{fig:similar_by_alfa_beta} 
\end{figure}

\noindent The values are time invariant, but depend on both $u$ and $v$. In Fig. \ref{fig:distrUSER_ITEM}, Fig. \ref{fig:BLCC1} and Fig. \ref{fig:similar_by_alfa_beta} three relations between model's parameters and graph's features are delineated.

\section{Experiments}
In order to evaluate the performance of analyzed algorithms we generated $83$ artificial bipartite graphs. The statistics describing the graphs are contained in Fig. \ref{fig:tables1} and Fig. \ref{fig:tables2}. Each graph's edge was augmented with a random integer from a set of possible rankings $\{0,1,2,3,4,5\}$. After the last iteration (usually $T=10~000$) hundred more edges were created by running 100 steps for each graph with unchanged parameters. This enabled us to preserve asymptotic properties of the graphs within a set of rankings used to batch update of the models. The experiments were run in-memory within separate threads on a 64-bit Fedora operating system with
the Quatro 2.66GHz Intel(R) Core(TM) i5 CPUs. 

\subsection{Evaluated systems}

We evaluated six recommender algorithms implemented in the Mahout java library \cite{mahout}. Mahout contains highly efficient open-source implementations of machine-learning algorithms maintained by a vibrant community. It is powering several portals e.g. \textit{SpeedDate}, \textit{Yahoo! Mail}, \textit{AOL} or \textit{Mippin}. The algorithms are: GenericUserBasedRecommender \cite{herlocker}, GenericItemBasedRecommender \cite{item}, SlopeOneRecommender \cite{slopeone}, GenericUserBasedRecommender with the neighborhood defined by non-negative threshold similarity \cite{mahoutinaction}, KnnItemBasedRecommender \cite{knn} and SVDRecommender \cite{svd}. The algorithms cover wide spectrum of approaches to the problems of \textit{Collaborative Filtering}. 

\subsection{Performance Measures}

In the subsequent subsections we study the correlations between the performance of the six algorithms differentiated by five graph properties: size, density, proportion of users to items, clustering and shape of node degree distributions. We focus our attention on four performance statistics:

\begin{enumerate}
	\item \textbf{BUILD} - time in milliseconds that is required to load whole bigraph from a text file and train a model, after this period of time the model is ready to create recommendations
	\item \textbf{MEMORY} - memory consumption in megabytes of the built model, it was assessed as a difference between \verb!Runtime.getRuntime().totalMemory()! and \verb!Runtime.getRuntime().freeMemory()! after calling a garbage collector five times
	\item \textbf{LATENCY} - an average time in milliseconds required to produce a recommendation for a sample of 500 users
	\item \textbf{UPDATE} - time in millisecond of updating a model with 100 new ratings
\end{enumerate}

In case of UserBased recommender we set the size of the neighborhood to $200$. SVD recommender was projected onto $10$ factors and $200$ iterations were run during training. The \textit{Pearson Correlation} was used as a similarity measure. At the end we analyze specific parameters of UserBased model.

\subsection{Scalability}

\begin{figure}[htbp]
\begin{tabular}{cc}
\centering
	a) BUILDING & b) MEMORY\\
\includegraphics[width=0.5\textwidth]{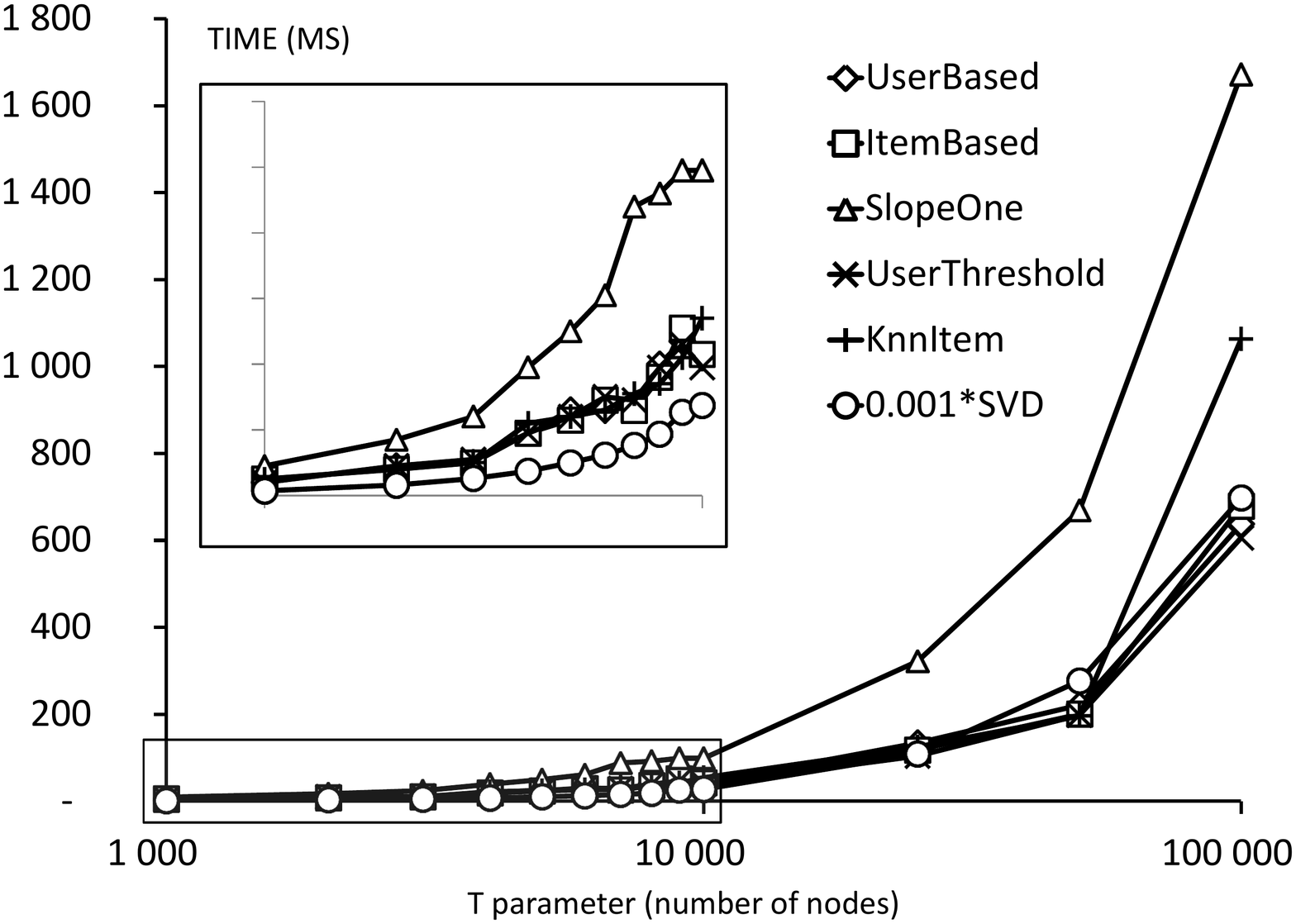}&\includegraphics[width=0.5\textwidth]{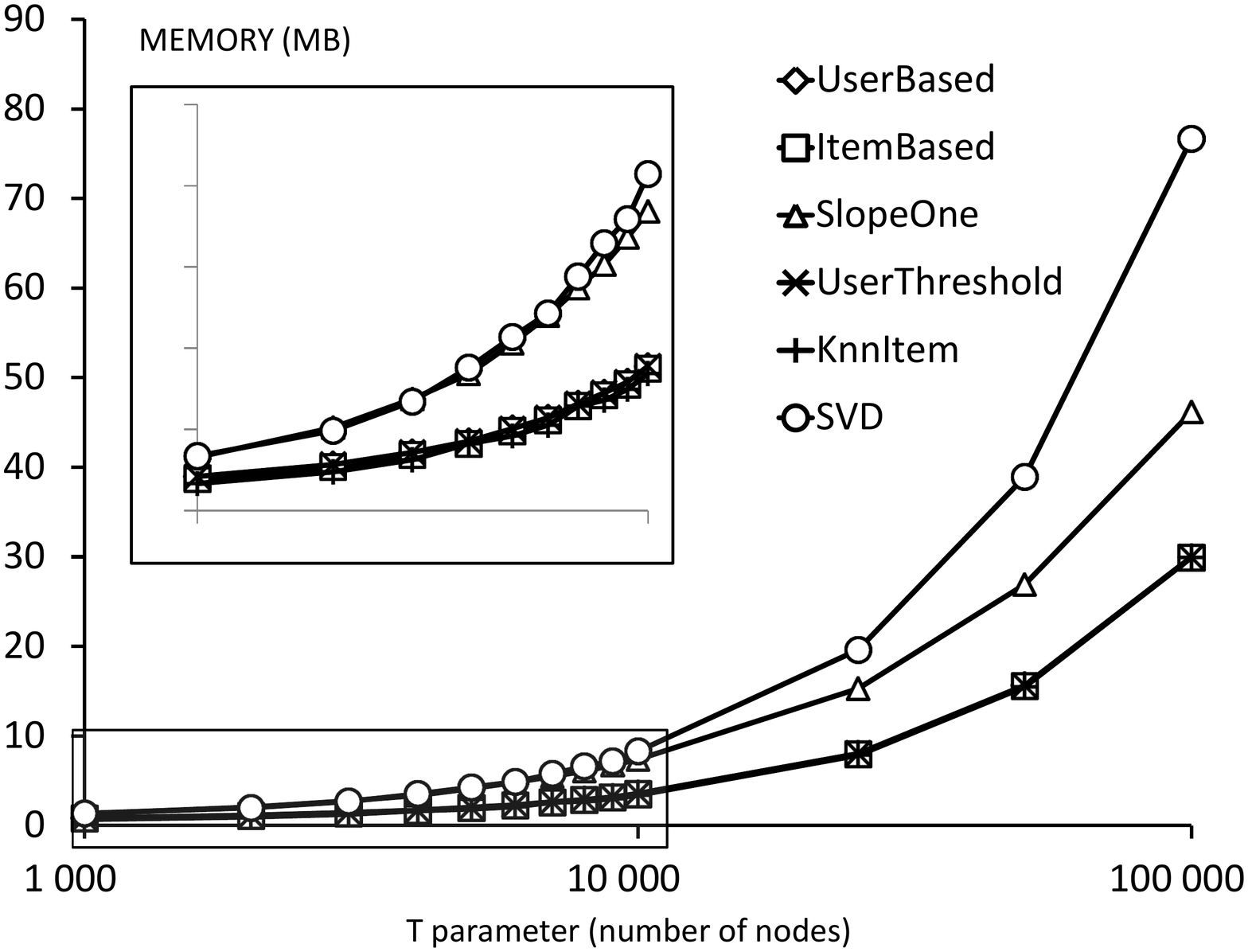}\\
c) LATENCY & d) UPDATING  \\
	\includegraphics[width=0.5\textwidth]{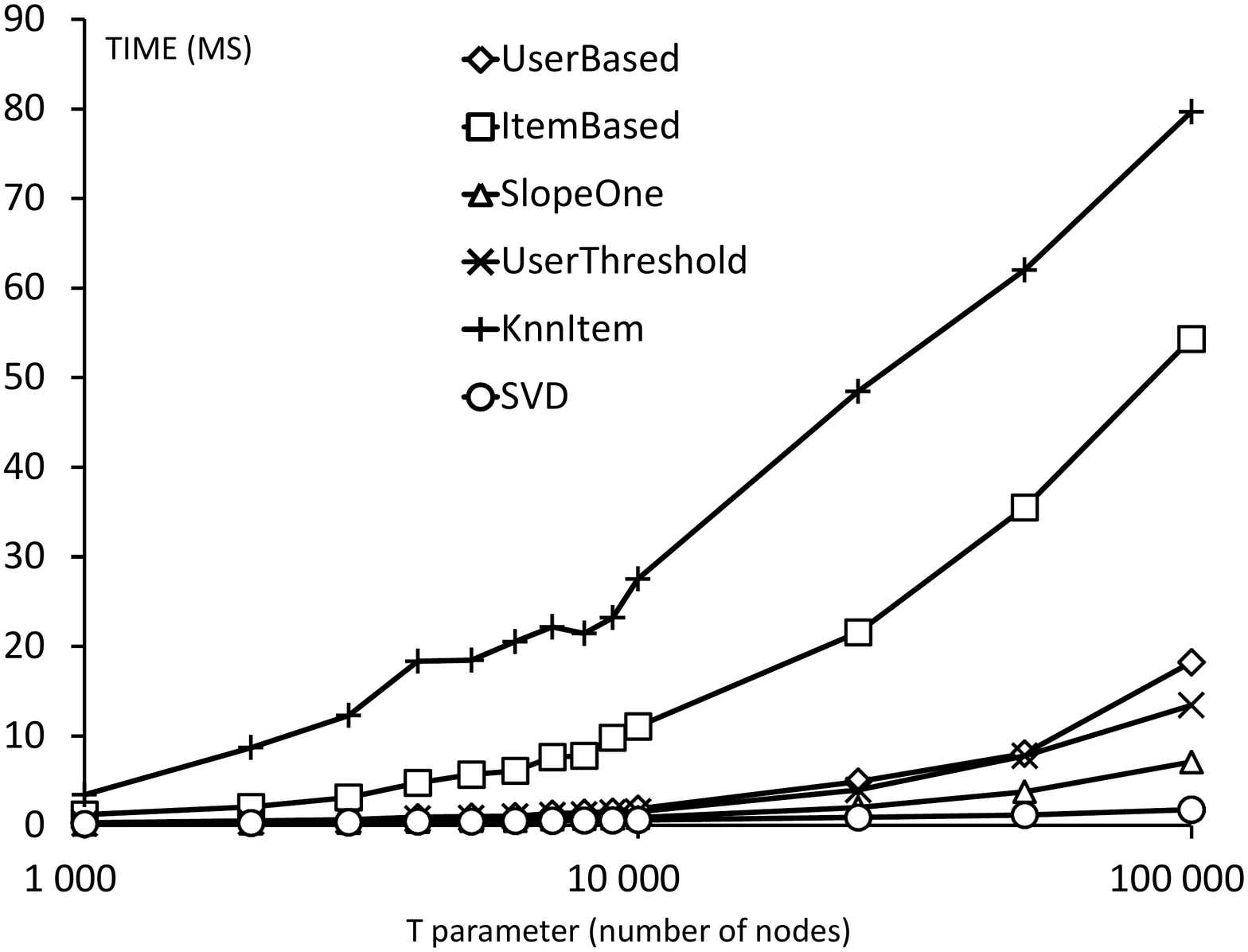}&
	\includegraphics[width=0.5\textwidth]{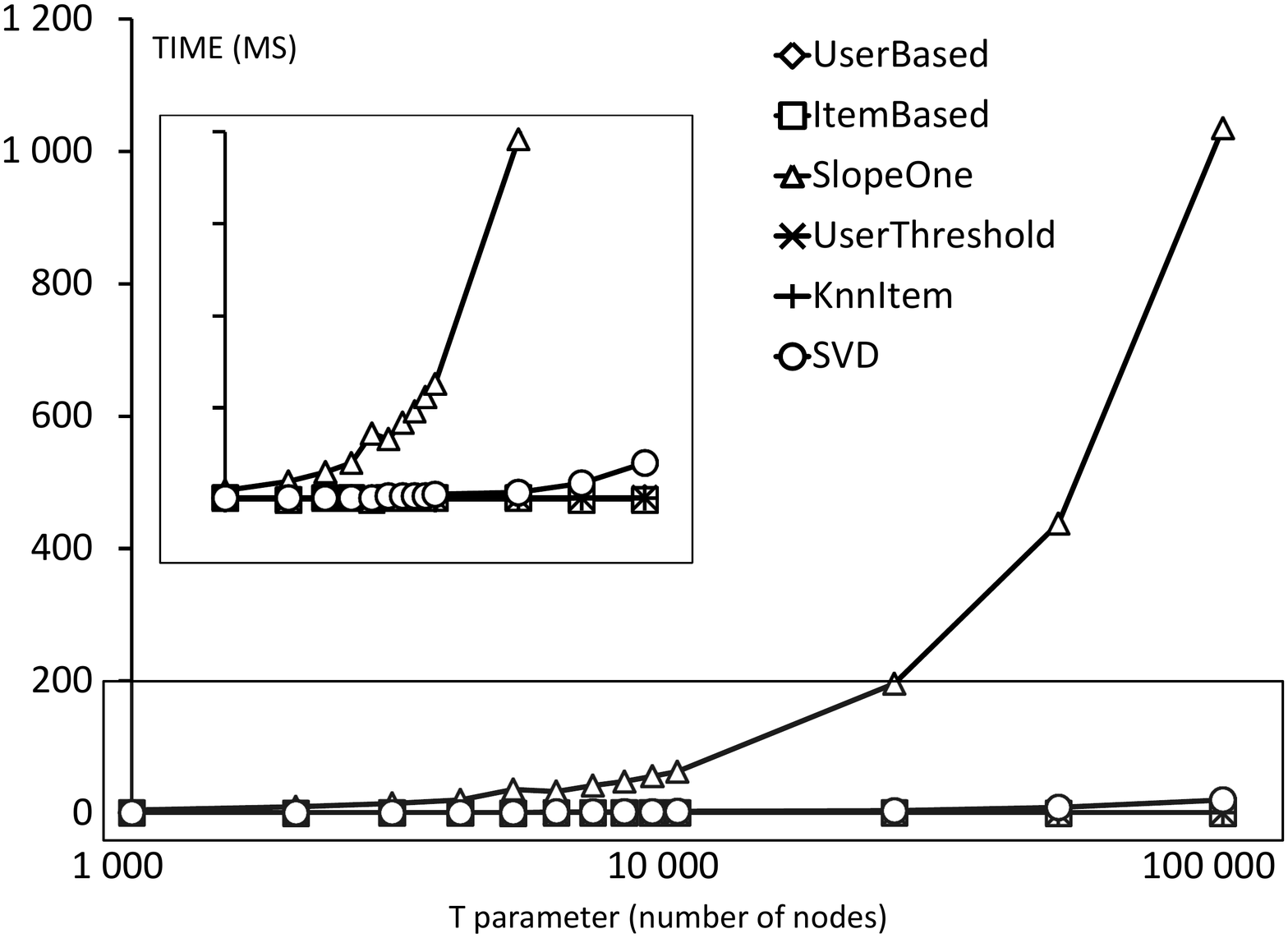}\\
	\end{tabular}
	\caption{Performance of the recommenders conditioned by the size of bigraphs. On all charts vertical axis has logarithmic scale. These enables to distribute evenly all observations.}
	\label{fig:scale}
\end{figure}

In order to verify the ability of analyzed algorithms to scale we generated thirteen bigraphs. Each dataset was produced with the same parameters except of the number of iterations. The datasets are numbered from 10 to 22 in Fig. \ref{fig:tables1}. The time of training SVD recommender may be misleading \footnote{In the chart (Fig. \ref{fig:scale}) the time of SVD building was divided by $1~000$ to get comparable results. Such long time of building the model is a result of running $200$ iterations of the gradient descent. However, even a single iteration takes on average $1~000/200=5$ times longer than building the other models.}. For most systems, only training time grows linearly with the size of a dataset. Memory consumption and latency grows sublinearly. SlopeOne and SVD algorithms exhibit the poorest performance in terms of building time and memory consumption. These costs pay back in the phase of creating recommendations. KnnItem does not scale during training and is the slowest during creating recommendations. Except of SlopeOne all models refresh their structures immediately. Only SVD's latency seems not to depend on the scale. 

\subsection{Density}

We generated 14 bigraphs to test how performance depends on the density. The graphs are numbered from 23 to 36 in Fig. \ref{fig:tables1}. All graphs have around the same number of nodes, but the number of edges varies between $30~100$ and $240~100$. The diversity was obtained by changing two parameters: $u$ and $v$. 

\begin{figure}[htbp]
\begin{tabular}{cc}
\centering		a) BUILDING & b) MEMORY\\
\includegraphics[width=0.5\textwidth]{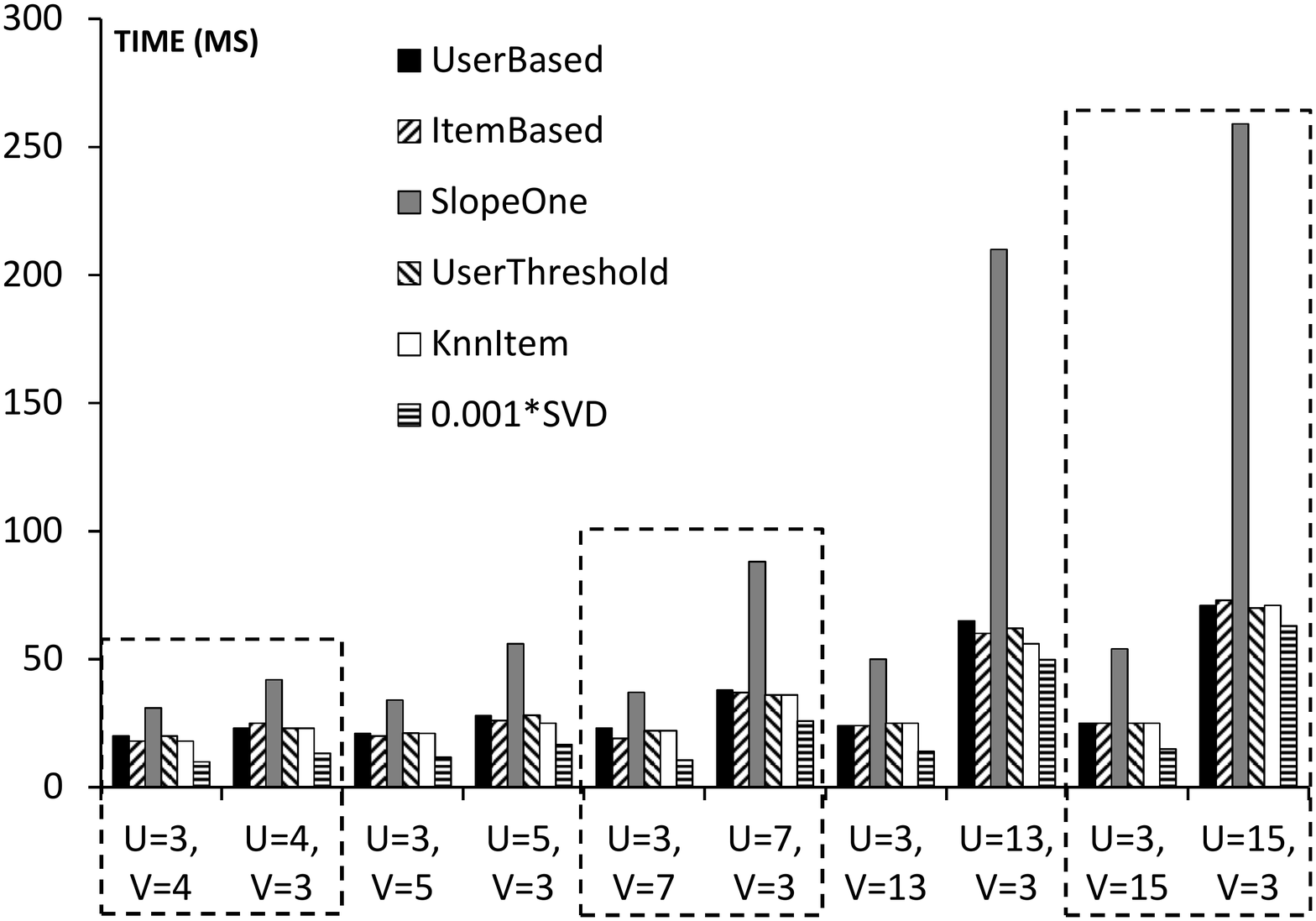}&\includegraphics[width=0.5\textwidth]{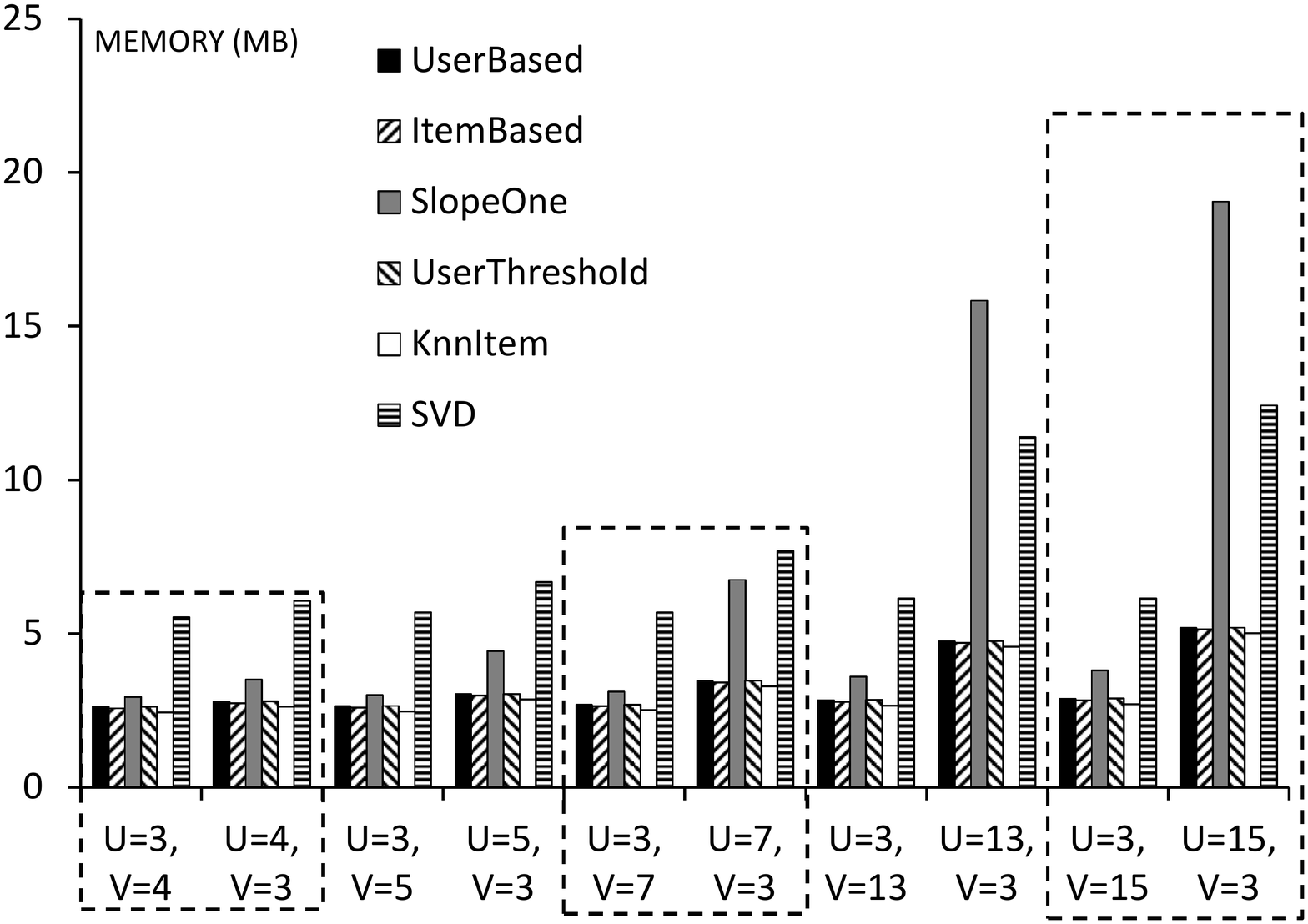}\\
c) LATENCY & d) UPDATING  \\
	\includegraphics[width=0.5\textwidth]{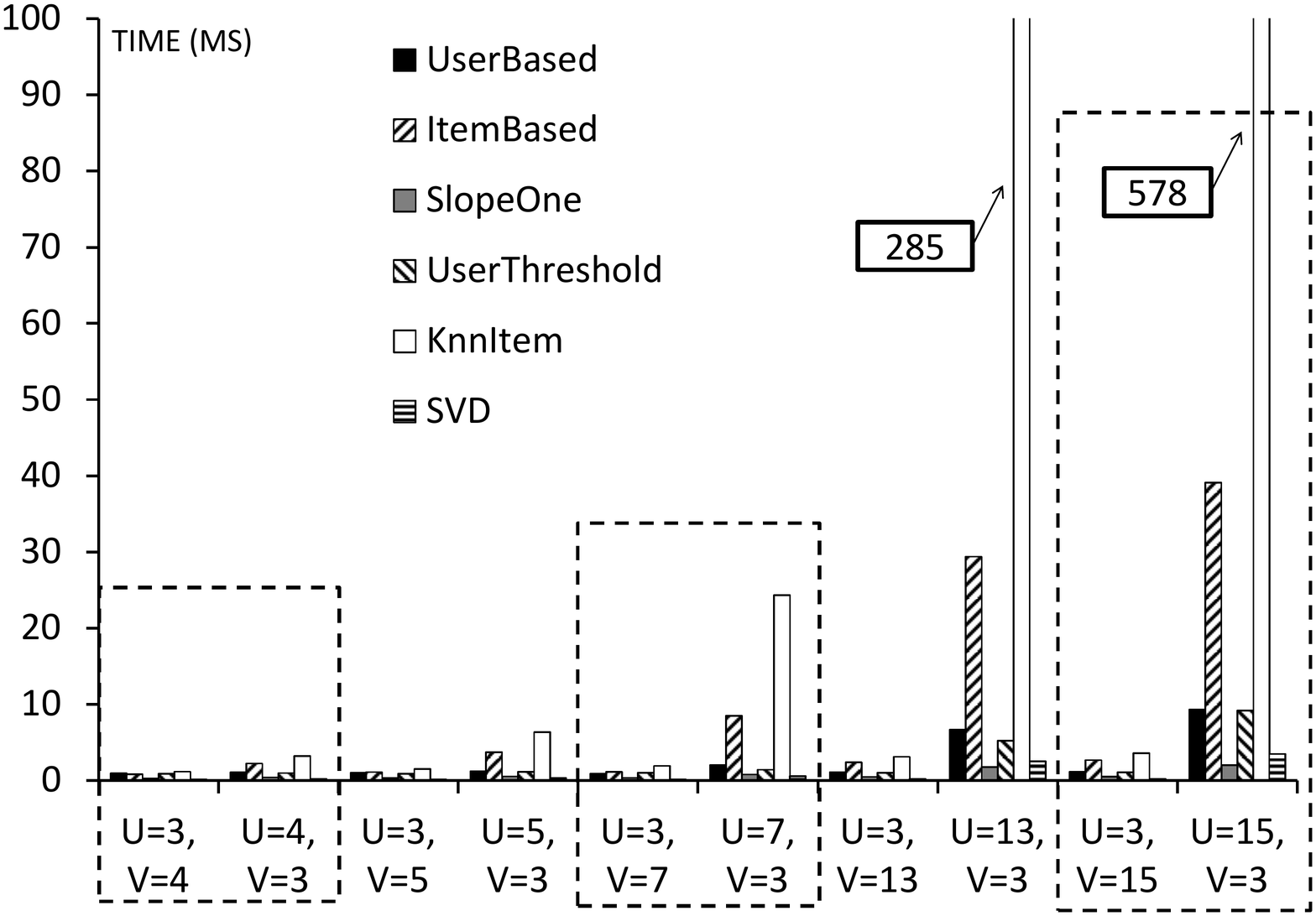}&
	\includegraphics[width=0.5\textwidth]{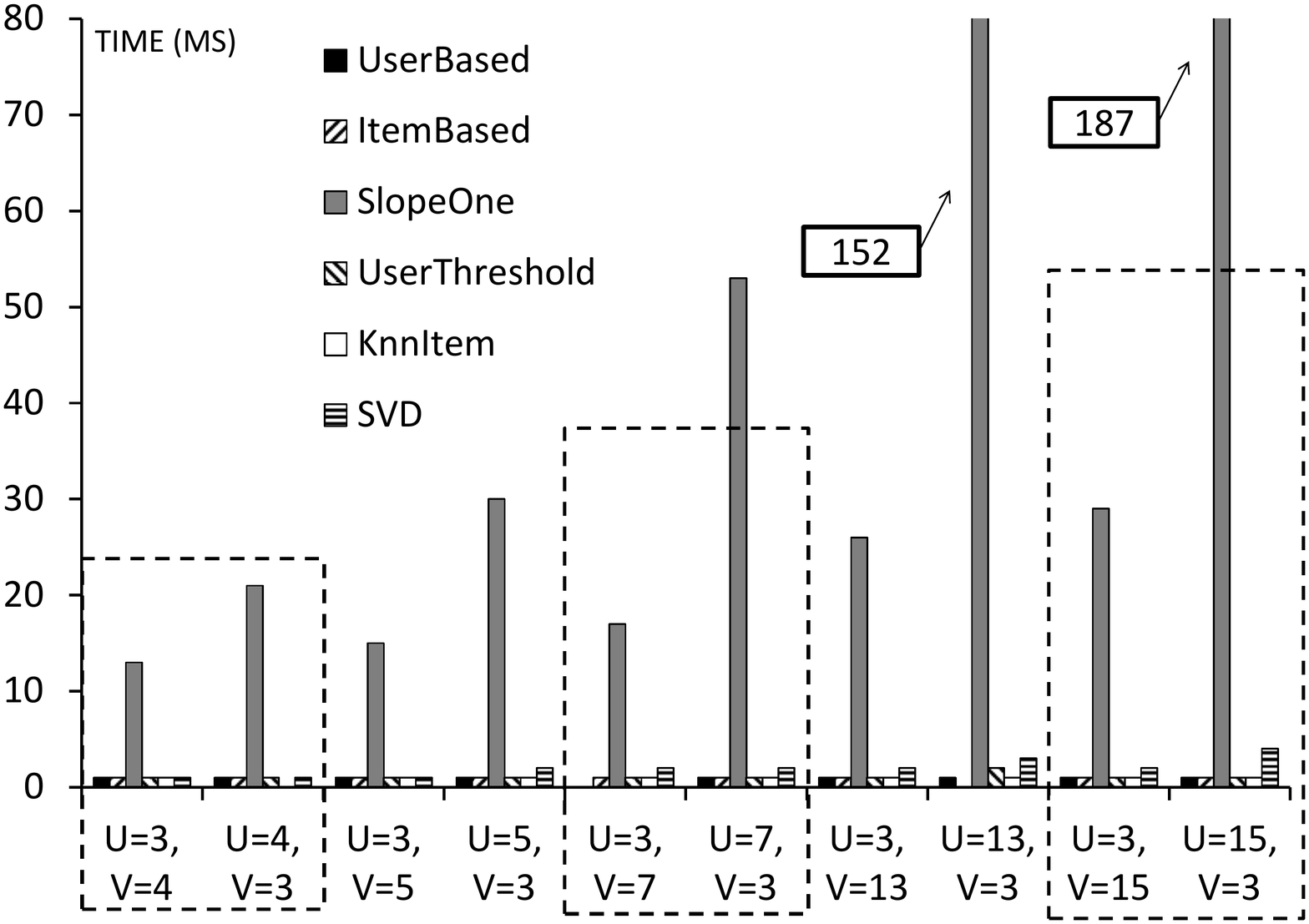}
\\
	\end{tabular}
	\caption{Performance of the recommenders versus the density of the graphs. }
	\label{fig:density}
\end{figure}

We omit the presentation of performance calculated for the first four bigraphs, in which $u=v$. All the four measures where growing steadily as we evaluated $(u=v=3)$, $(u=v=6)$, $(u=v=12)$ and $(u=v=24)$. Which could have suggested that there exists a strong correlation between the density and the performance. However, when we compare 5 pairs of graphs with $u\neq v$ (Fig. \ref{fig:density}), the results become confusing. By setting $u=3$ and iterating over $\{4,\dots,15\}$ with $v$ the performance diminishes only slightly. If we freeze $v=3$ and iterate with $u$ we decrease the performance steadily. It suggests that the density on its own is not so much responsible for the performance\footnote{Digging this effect deeper we see that by increasing $v$ we not only increase the density, but also lower the variance of node degree distribution in user modality stronger than for the item modality. This shows that the performance relies heavily on the distributions of node degrees, but with different magnitude for both modalities.}.

\subsection{Users to Items Proportion}

Parameter $p$ in the generator enables us to control the proportion of users to items. The parameter is interpreted as the probability that a new node is a user. We built nine graphs with constant numbers of nodes ($=10~200$) and edges ($=70~100$). The number of users varies between $1~166$ and $9~082$. The graphs are the first nine graphs in Fig. \ref{fig:tables1}.

\begin{figure}[htbp]
\begin{tabular}{cc}
\centering
	a) BUILDING & b) MEMORY\\
\includegraphics[width=0.5\textwidth]{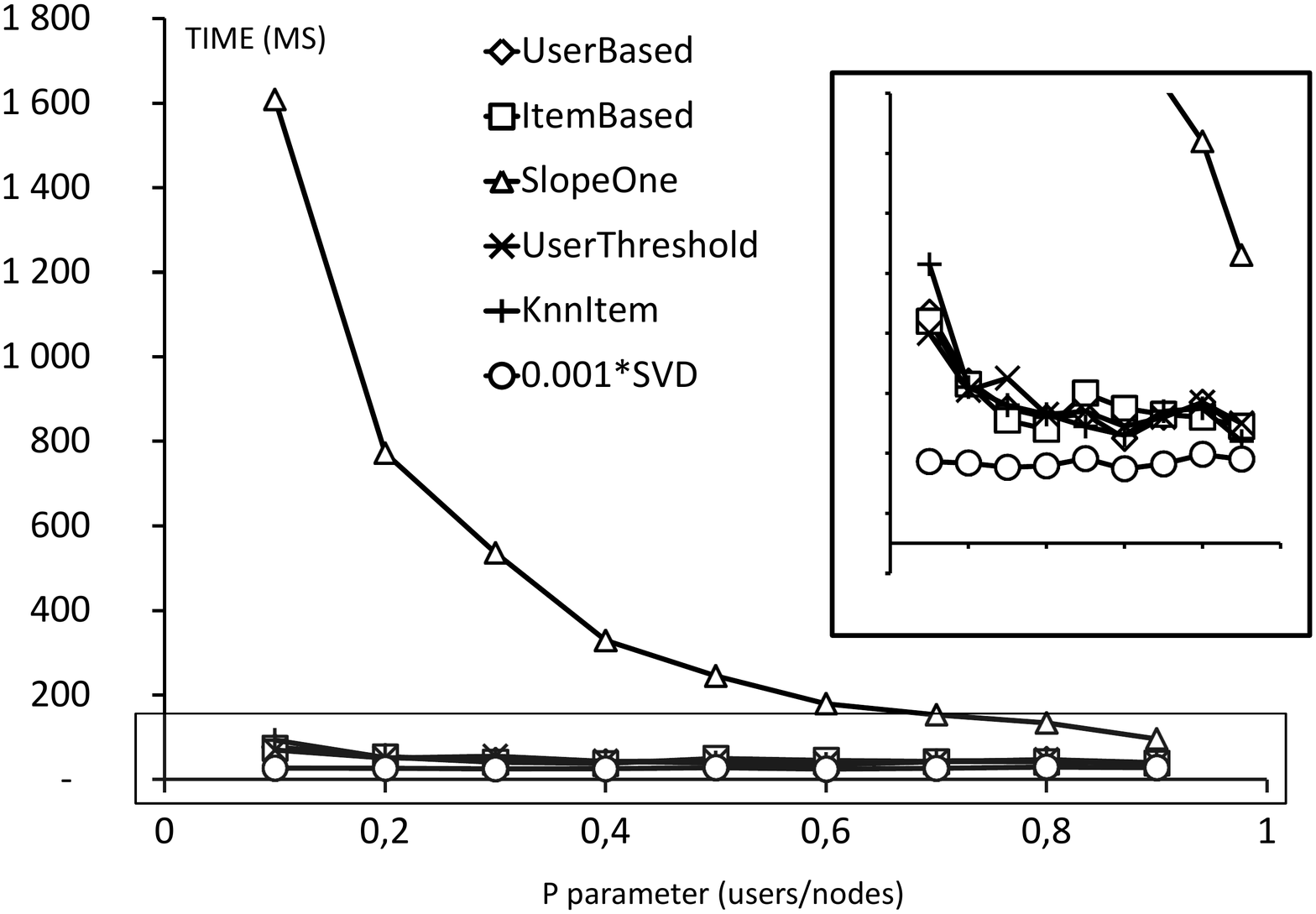}&\includegraphics[width=0.5\textwidth]{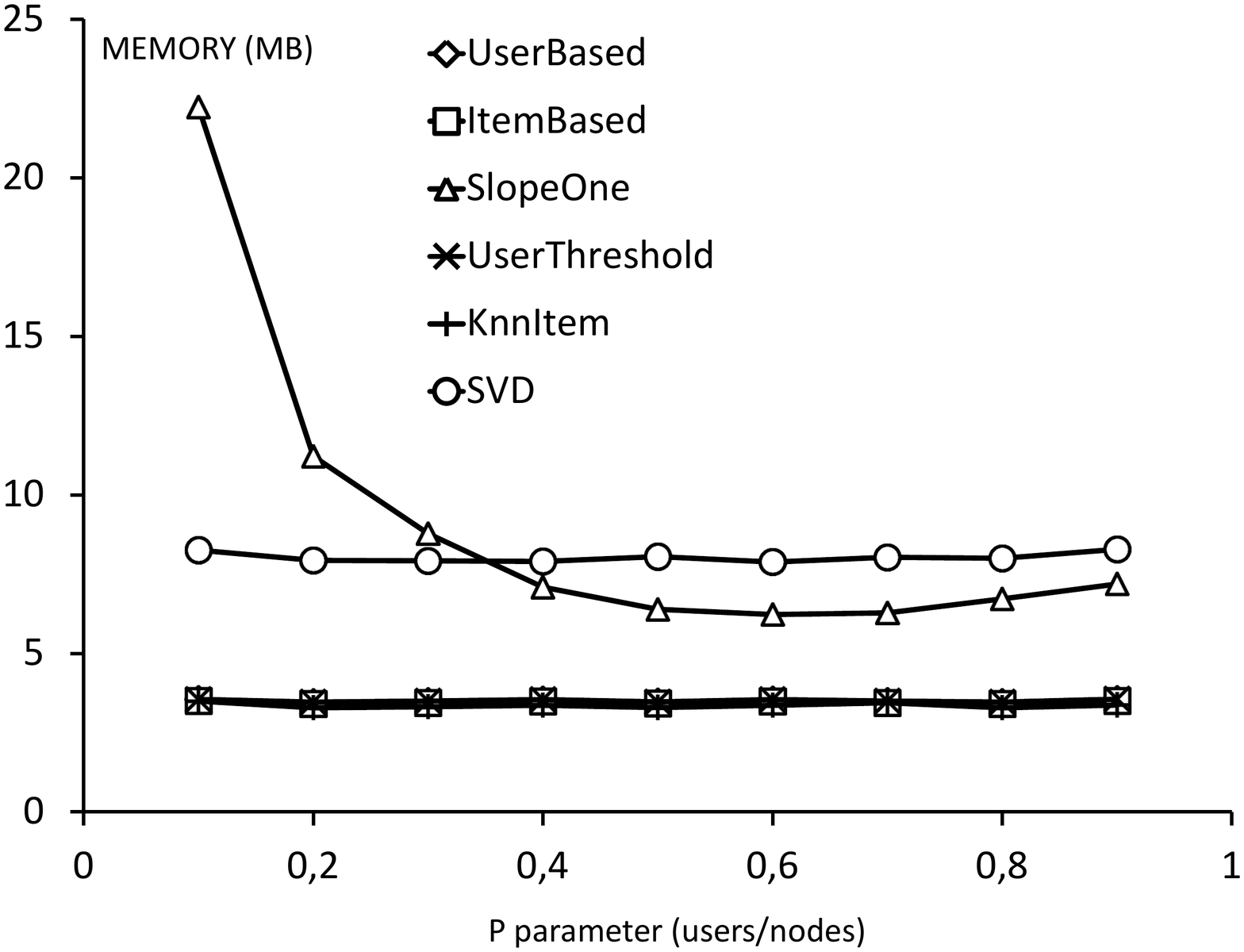}\\
c) LATENCY & d) UPDATING  \\
	\includegraphics[width=0.5\textwidth]{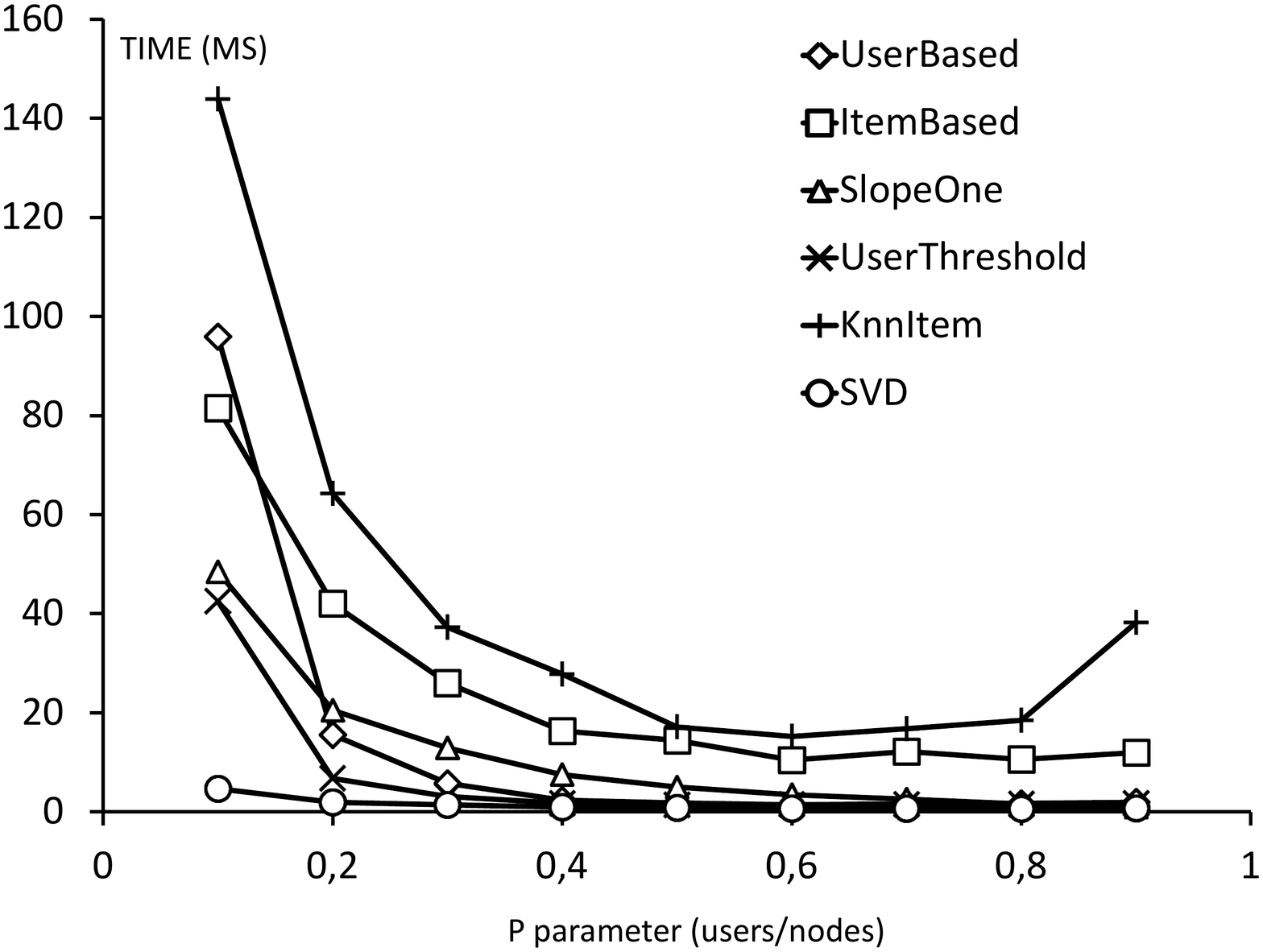}&
	\includegraphics[width=0.5\textwidth]{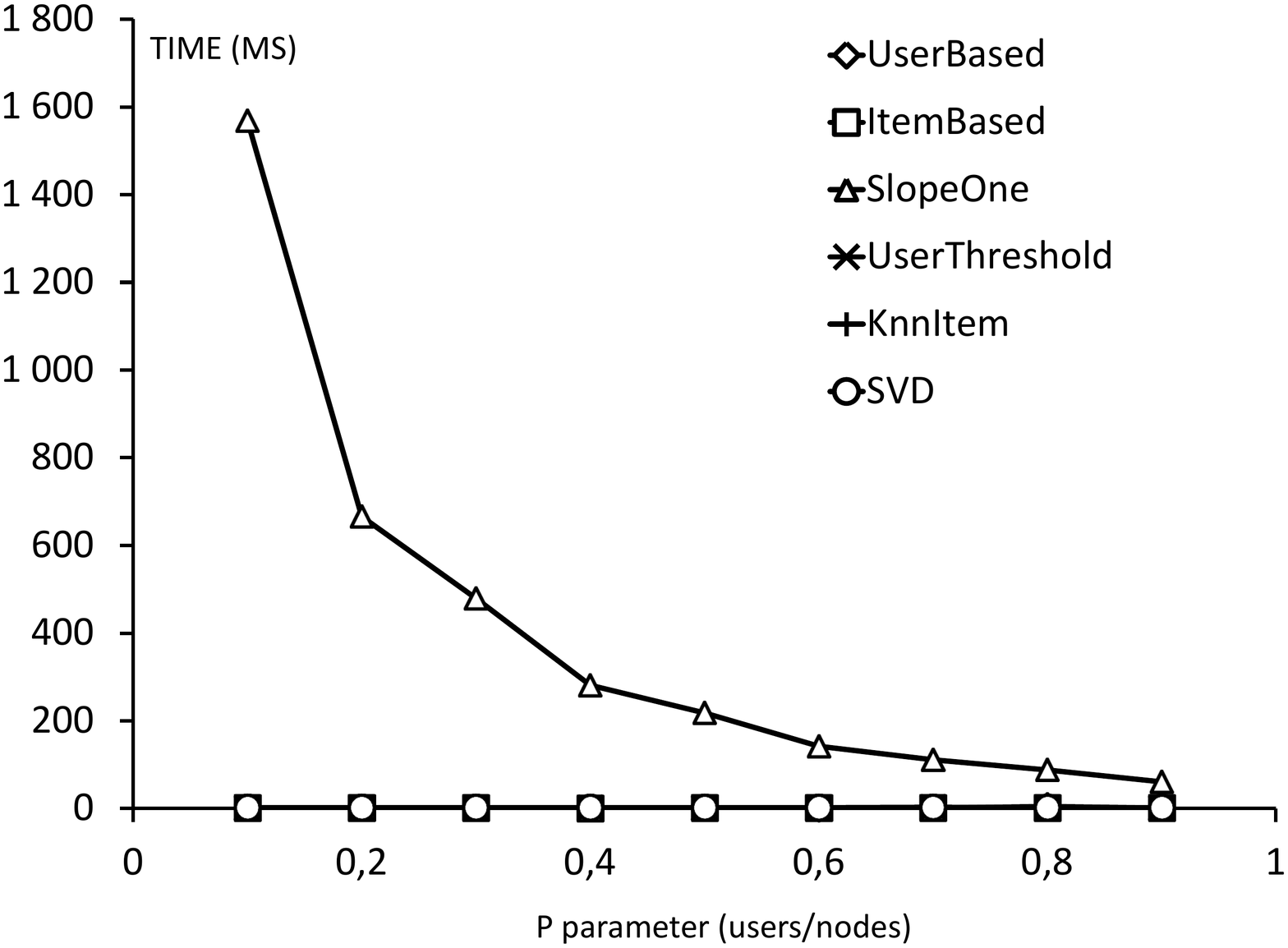}\\
	\end{tabular}
	\caption{Performance modeled by the proportion of users to all nodes. }
	\label{fig:proportion}
\end{figure}

The relationship between $p$ and the performance is depicted in Fig. \ref{fig:proportion}. Only the performance of SVD does not seem to depend on the proportion in none of four evaluations. Training time decreases as the relative number of users increases for all remaining models. The only model, which memory consumption depends on the proportion is SlopeOne. It is very interesting that the lowest memory requirements are obtained for balanced graphs ($p\approx 0.5$). The same non-monotonic relation can be observed in case of KnnItem and latency. The rest of the algorithms improve their latency as the relative number of items grows. Even though most datasets have more items than users, the opposite situations can also happen. For example there are around $500K$ questions in \url{http://stackoverflow.com/} technical portal and only around $200K$ users. The fact that ItemBased recommender does not perform better than UserBased when there are relatively few items (\cite{mahoutinaction}) raises our concern\footnote{This observation requires further investigation. In particular, we plan to verify if the claim that \textit{UserBased algorithm are preferred over ItemBased when there are fewer users} is valid only when caching mechanisms are implemented.}.

\subsection{Clustering}

\begin{figure}[htbp]
		\includegraphics[width=\textwidth]{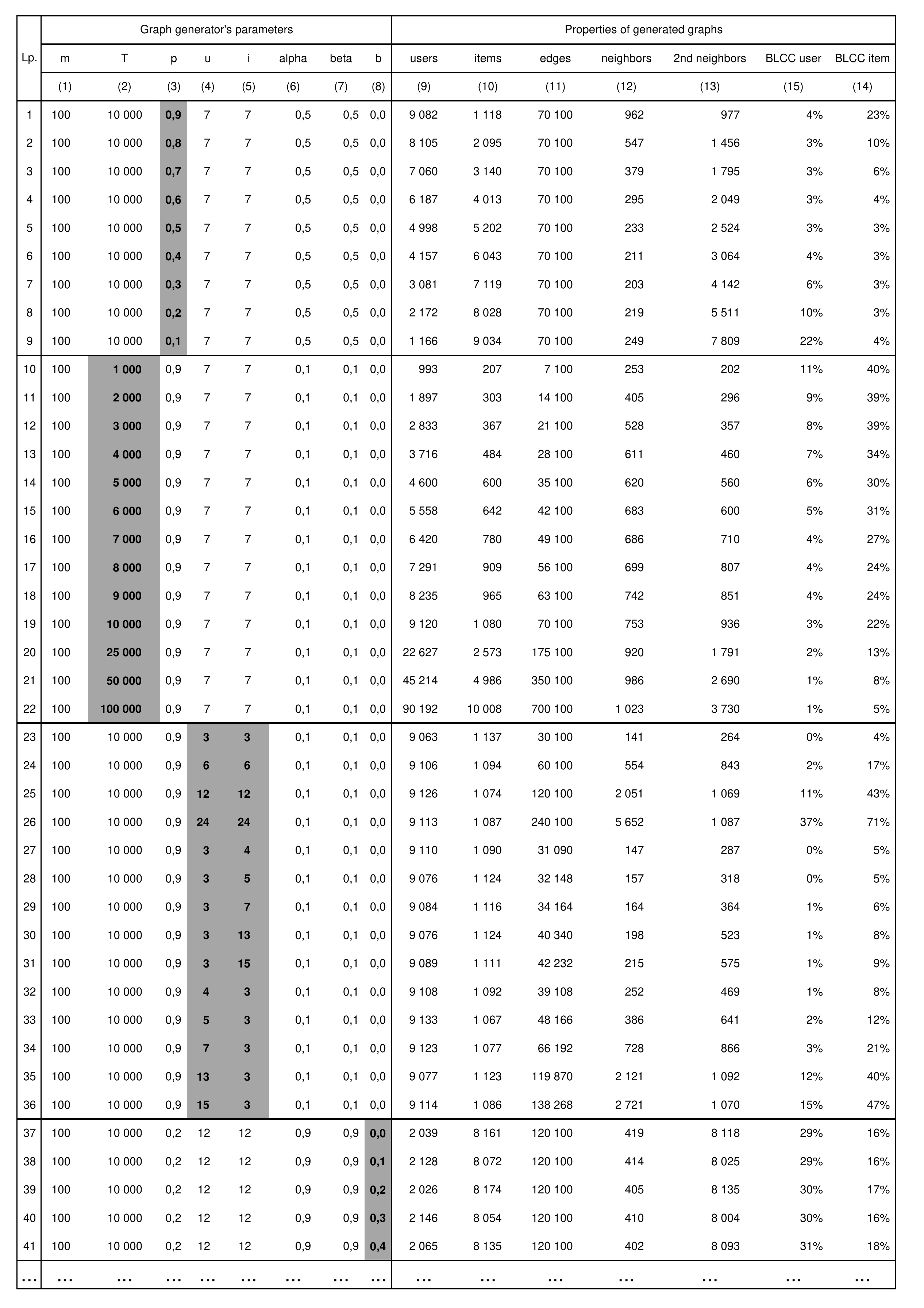}
		\caption{Random graphs generated to test the performance of recommender systems (part 1). Graph generator's parameters and BLCC are defined in Sec 3.1. Neighbors stands for an average number of potentially similar users, $2^{nd}$ neighbors stands for an average number of items of potentially simialar users. }
		\label{fig:tables1}
\end{figure}

Eleven graphs were generated to measure the influence of clustering on the performance. The datasets are numbered from $37$ to $47$ in Fig. \ref{fig:tables1} and Fig. \ref{fig:tables2}. The bigraphs were generated by changing the bouncing parameter $b$ from $0.0$ to $1.0$ with $0.1$ intervals. All graphs have $1~100$ nodes and $120~100$ edges.

The performance of SVD does not depend on bouncing. Memory consumption and time of building grow slightly in line with clustering only in case of SlopeOne model. The latency of creating a recommendation is correlate with clustering gently only in case of ItemBase, UserBase, KnnItem and UserThreshold algorithms. The relation between the clustering and the performance is in all variants weak and not stable.

\begin{figure}[htbp]
\begin{tabular}{cc}
\centering
	a) BUILDING & b) MEMORY\\
\includegraphics[width=0.5\textwidth]{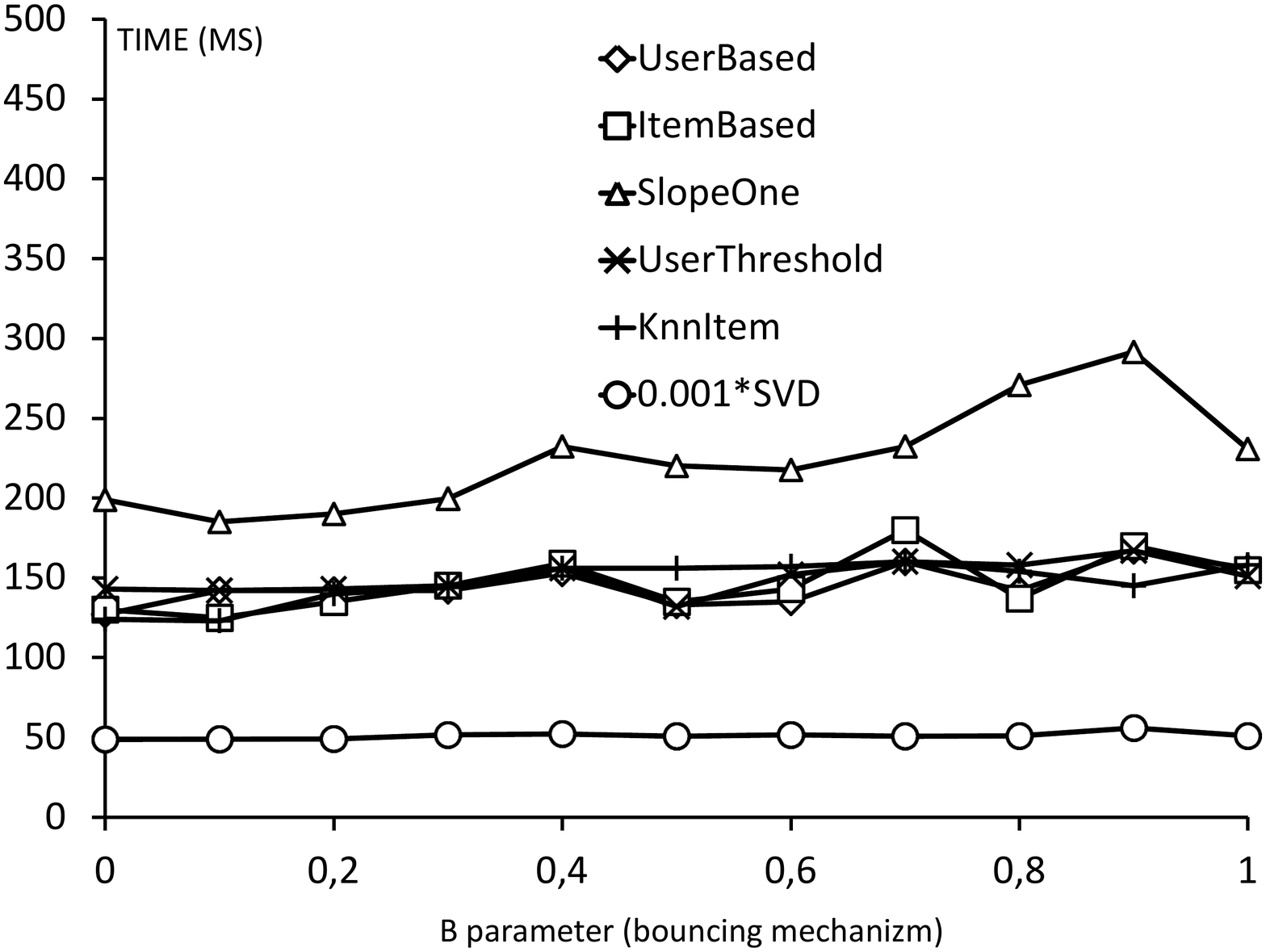}&\includegraphics[width=0.5\textwidth]{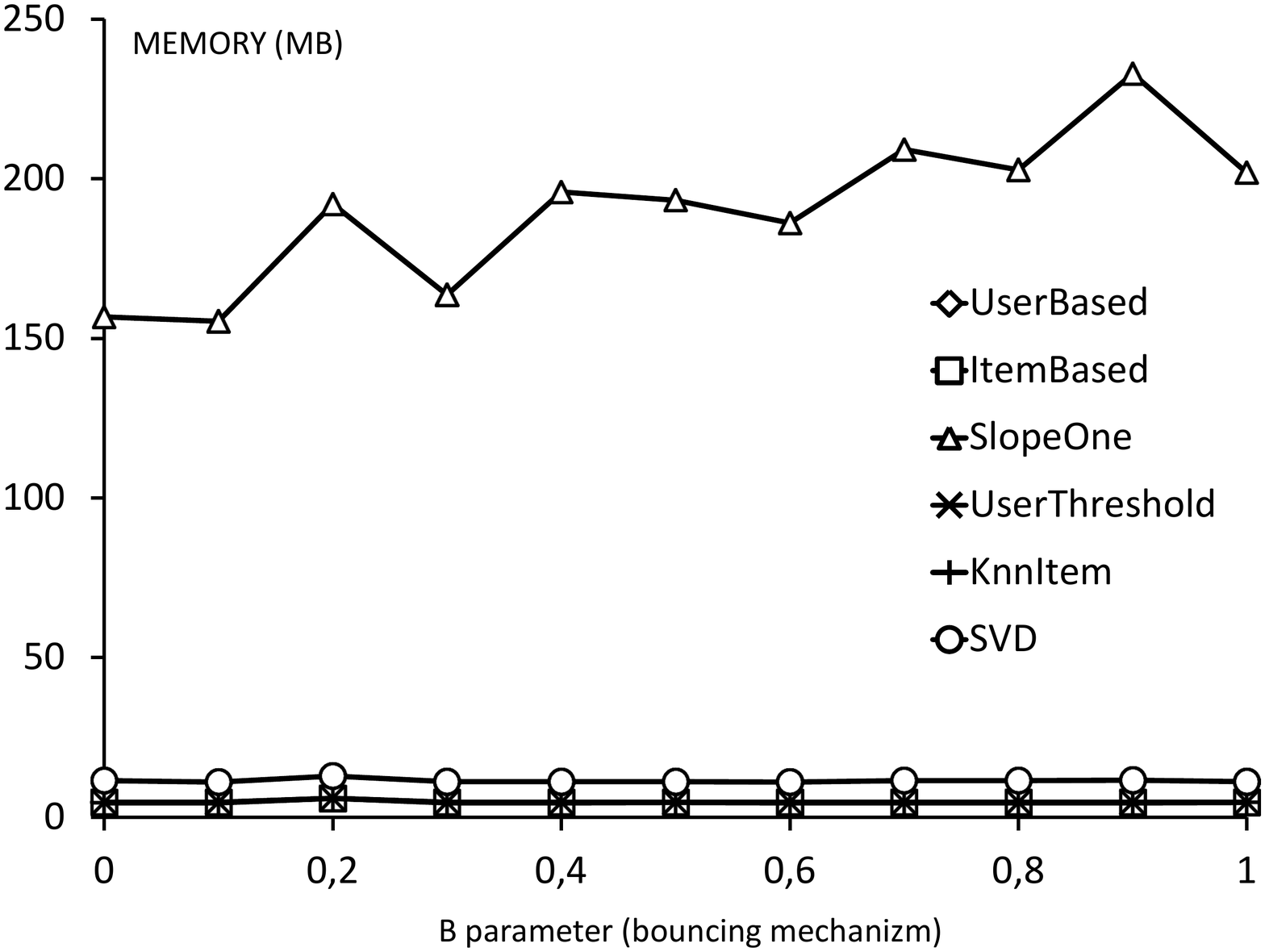}\\
c) LATENCY & d) UPDATING  \\
	\includegraphics[width=0.5\textwidth]{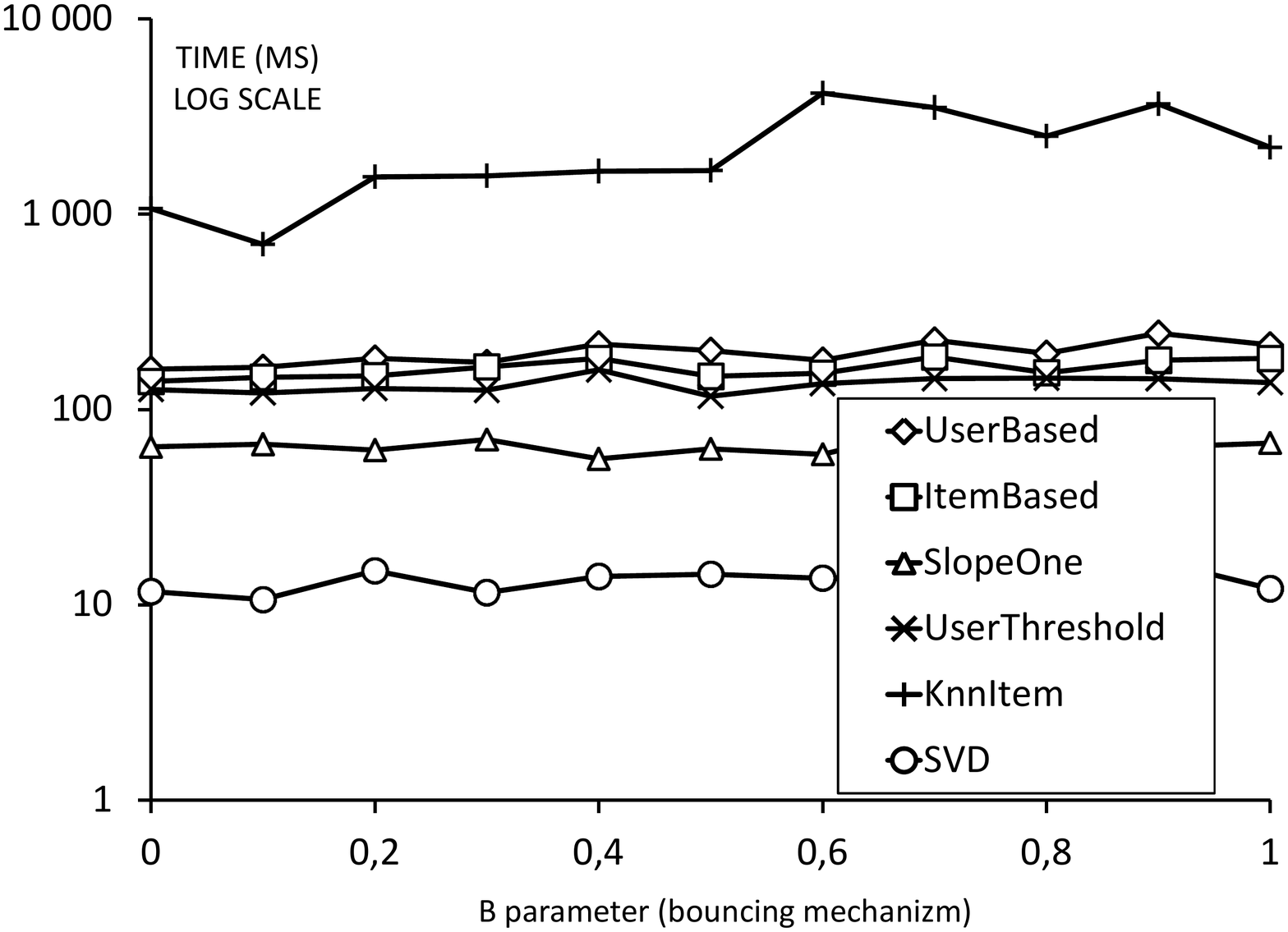}&
	\includegraphics[width=0.5\textwidth]{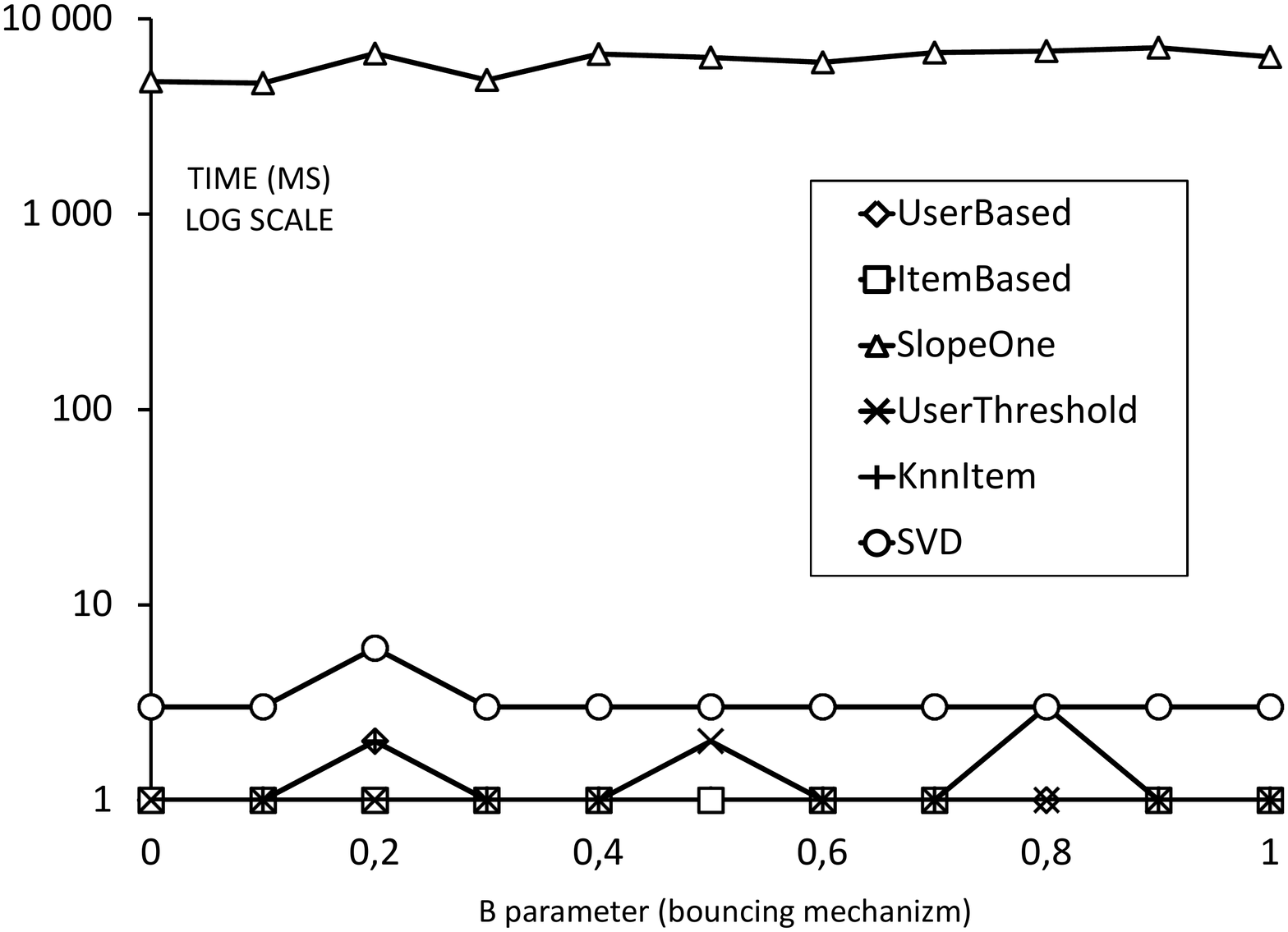}
\\
	\end{tabular}
	\caption{Performance dimensioned by various levels of clustering (transitivity).}
	\label{fig:BLCC_by_bounce}
\end{figure}

\newpage
\subsection{Shapes of degree distributions}

We have shown in \cite{chojnacki_CC} that by changing $\alpha$ and $\beta$ we can output graphs with a mixture of the power-law distribution and the exponential distribution (compare Fig. \ref{fig:distrUSER_ITEM}). The two parameters enable us to control independently the shapes of degree distributions of both modalities. By increasing $\alpha$ we can obtain graphs with constant number of users, items and edges, but growing average number of potentially similar users\footnote{We say that two users to be potentially similar if they have rated at least one item in common.}. Moreover, an average number of distinct items of potentially similar users grows with both $\alpha$ and $\beta$ (compare Fig. \ref{fig:similar_by_alfa_beta}). These results are consistent with an asymptotic Newman's formula \cite{Newman_2001}. The formula is based on a local tree-like structure assumption and assess as an average number of second neighbors $|N_2(j)|$ of a random user $j$ by means of the first moment of user degree distribution $\langle U \rangle$ and the first and the second moments of item degree distribution ($\langle I^2 \rangle$ and $\langle I^2 \rangle$ respectively):

\begin{equation}
|N_2(j)|=\langle U \rangle \left(\frac{\langle I^2 \rangle}{\langle I \rangle} -1 \right).
\end{equation}

\begin{figure}[htbp]
\begin{tabular}{cc}
	
	\includegraphics[width=0.5\textwidth]{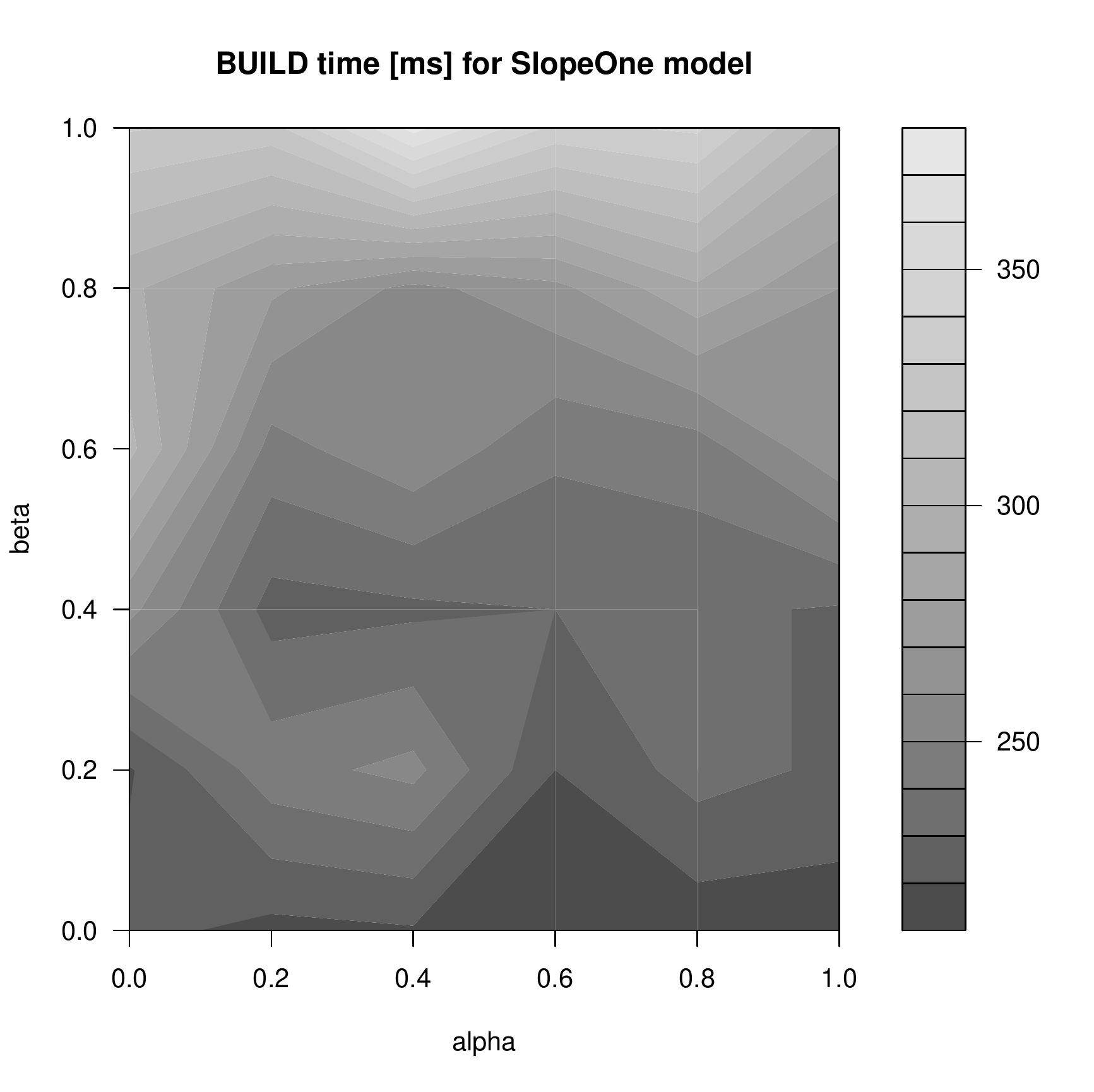}&\includegraphics[width=0.5\textwidth]{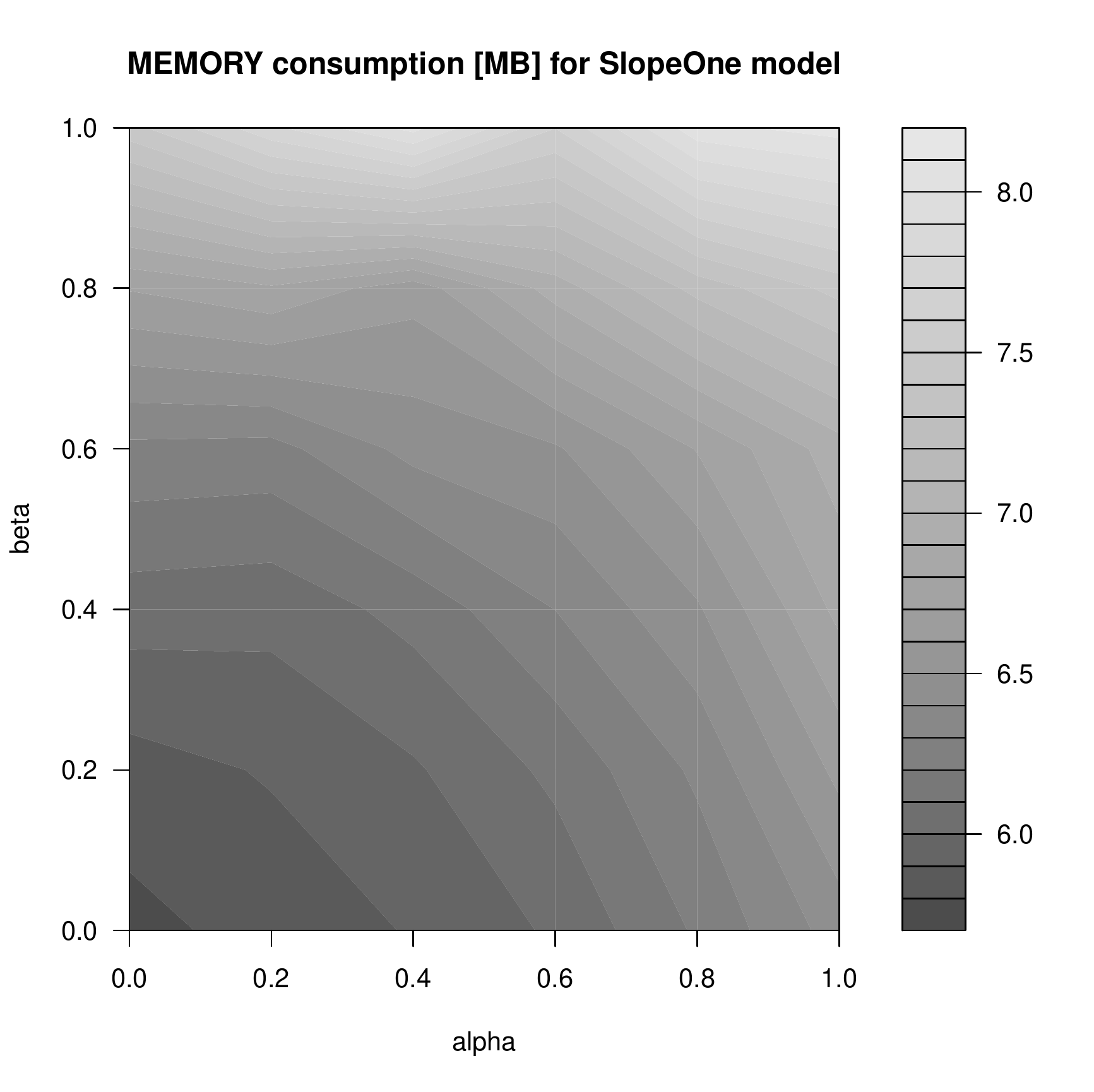}\\
	\end{tabular}
	\caption{Training time and memory requirements of SlopeOne model.}
	\label{fig:albe1}
\end{figure}

\begin{figure}[htbp]
\begin{tabular}{cc}
	
	\includegraphics[width=0.5\textwidth]{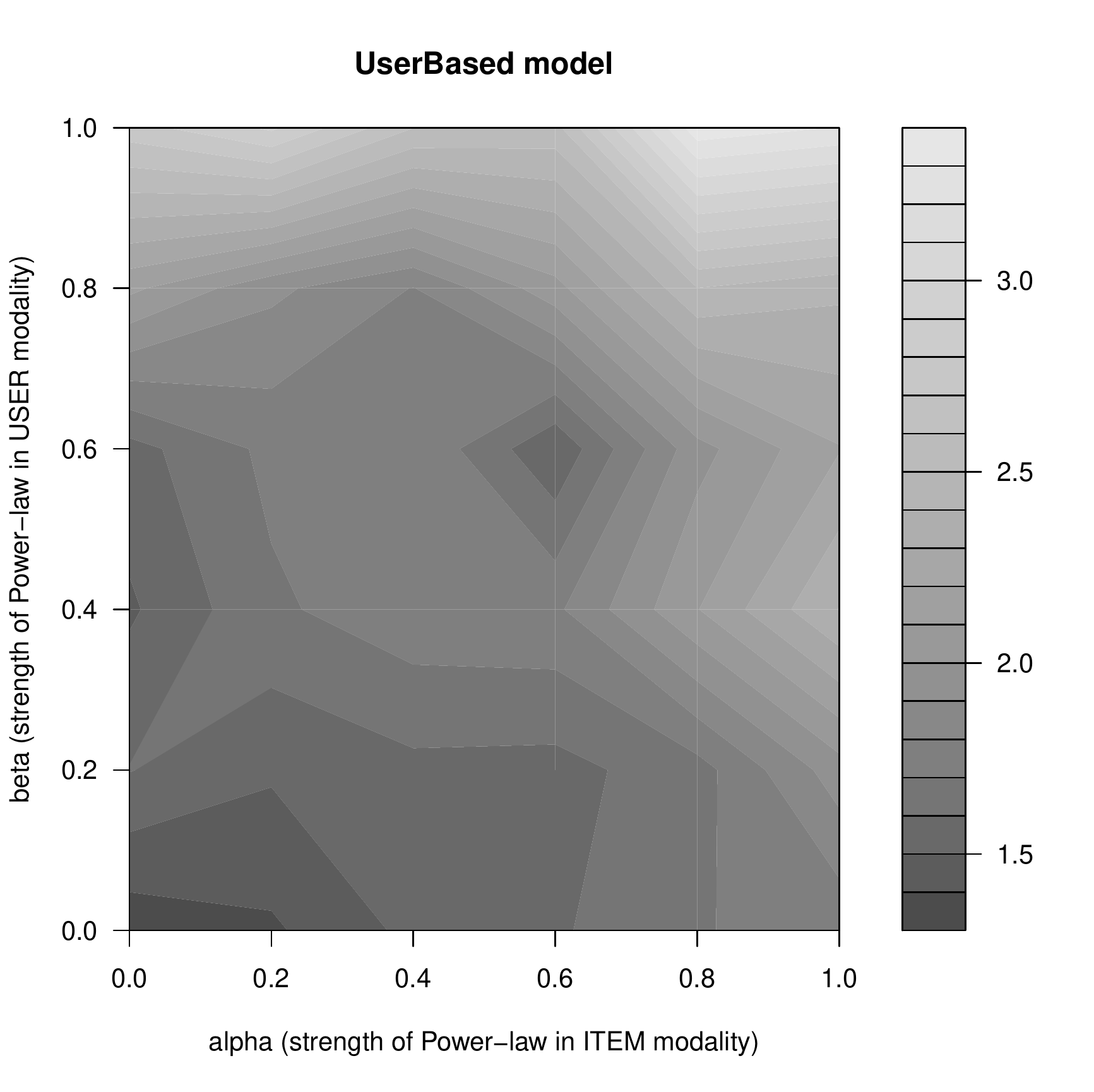}&\includegraphics[width=0.5\textwidth]{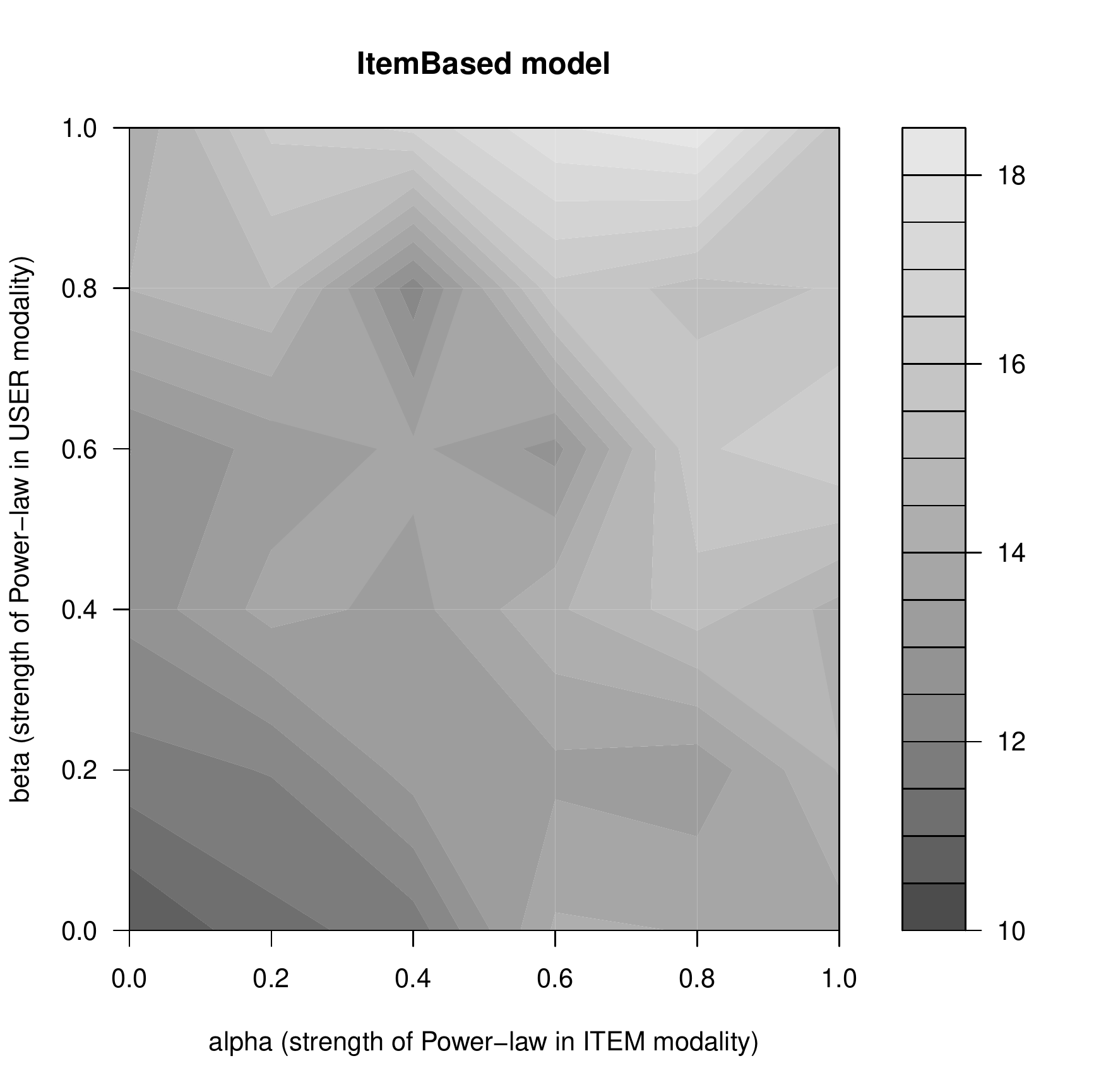}\\
	\includegraphics[width=0.5\textwidth]{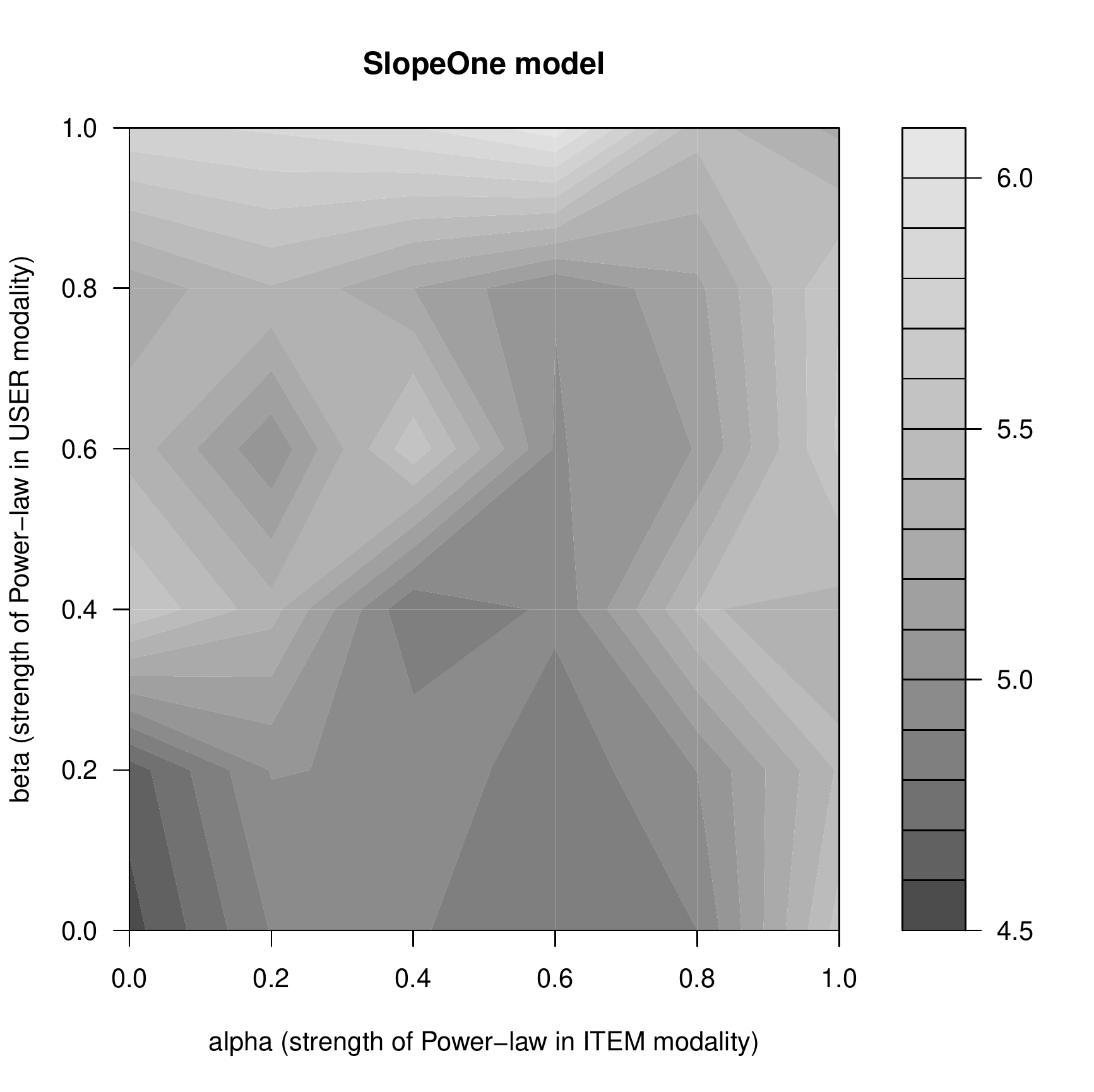}&\includegraphics[width=0.5\textwidth]{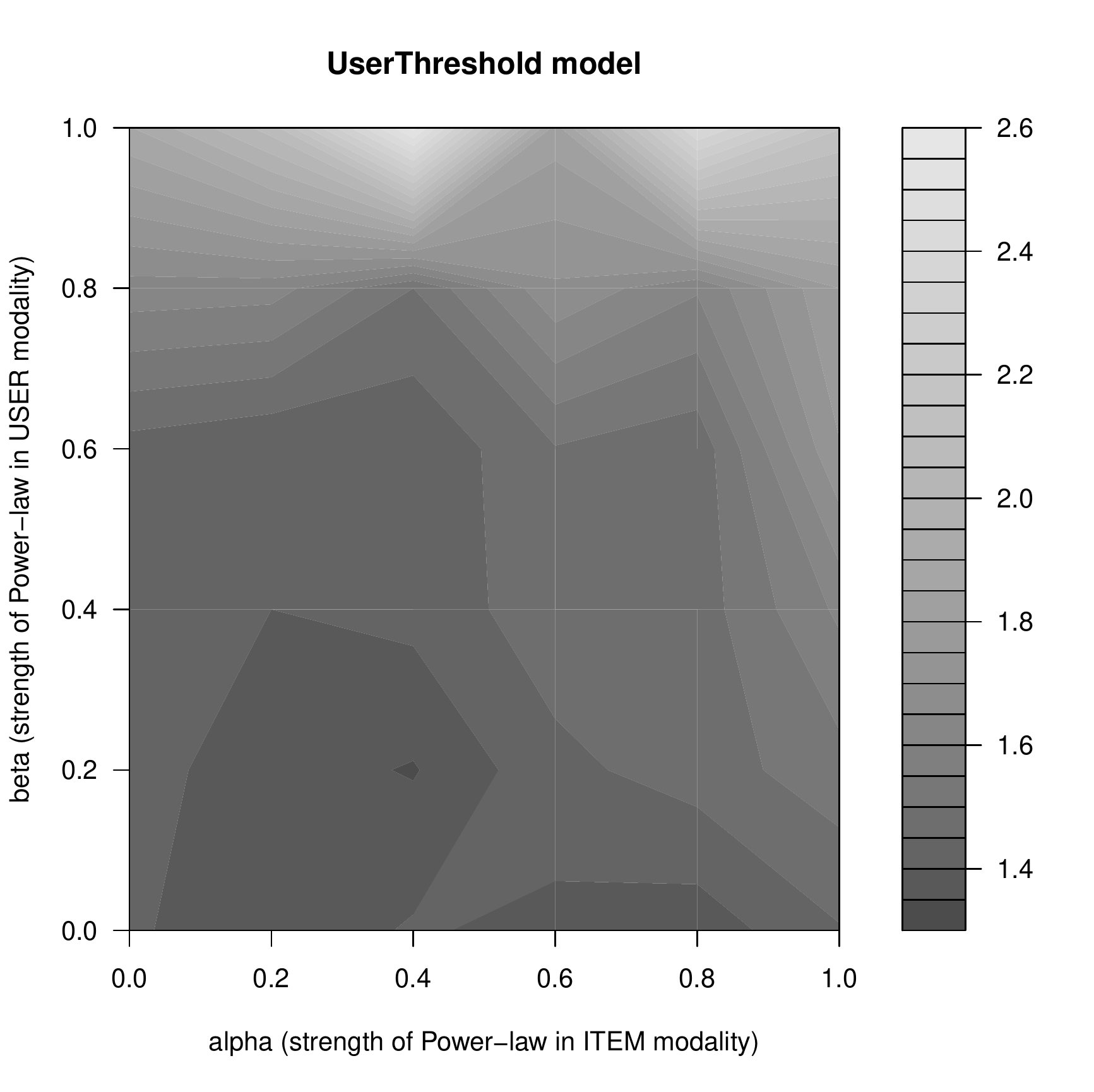}\\
	\includegraphics[width=0.5\textwidth]{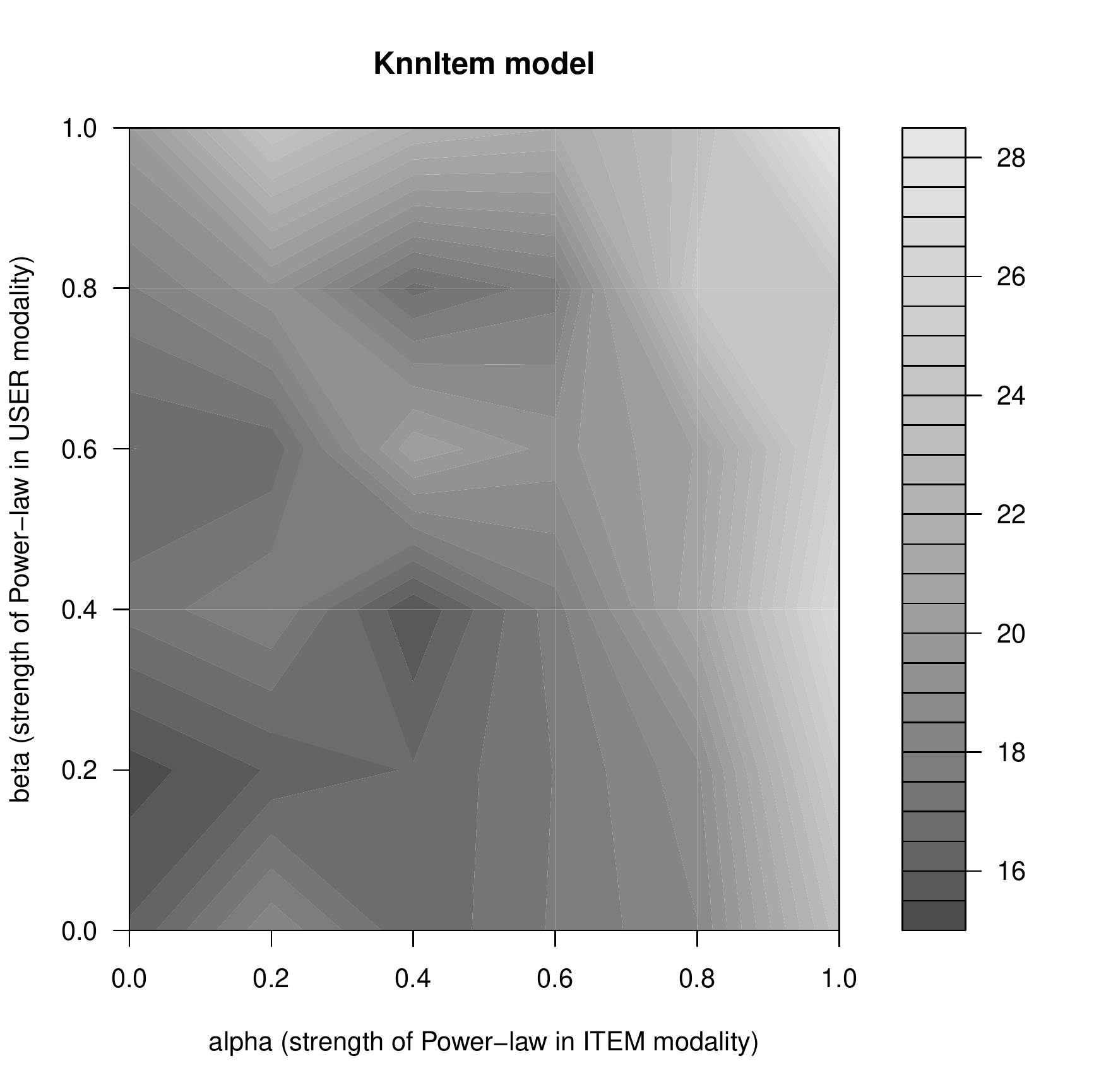}&\includegraphics[width=0.5\textwidth]{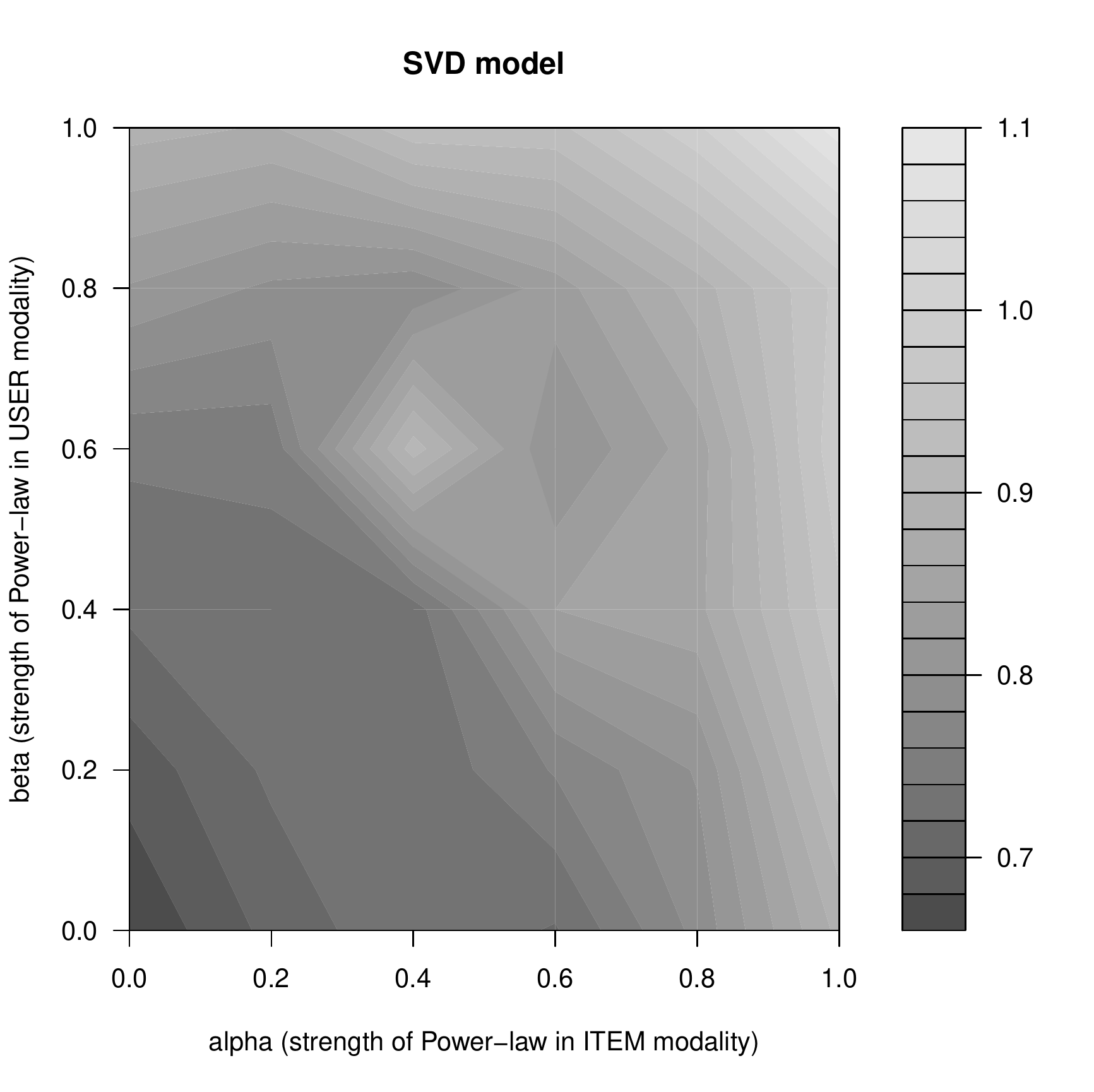}\\

	\end{tabular}
	\caption{The influence of $\alpha$ and $\beta$ parameters on the latency.}
	\label{fig:albe2}
\end{figure}

\begin{figure}[htbp]
		\includegraphics[width=\textwidth]{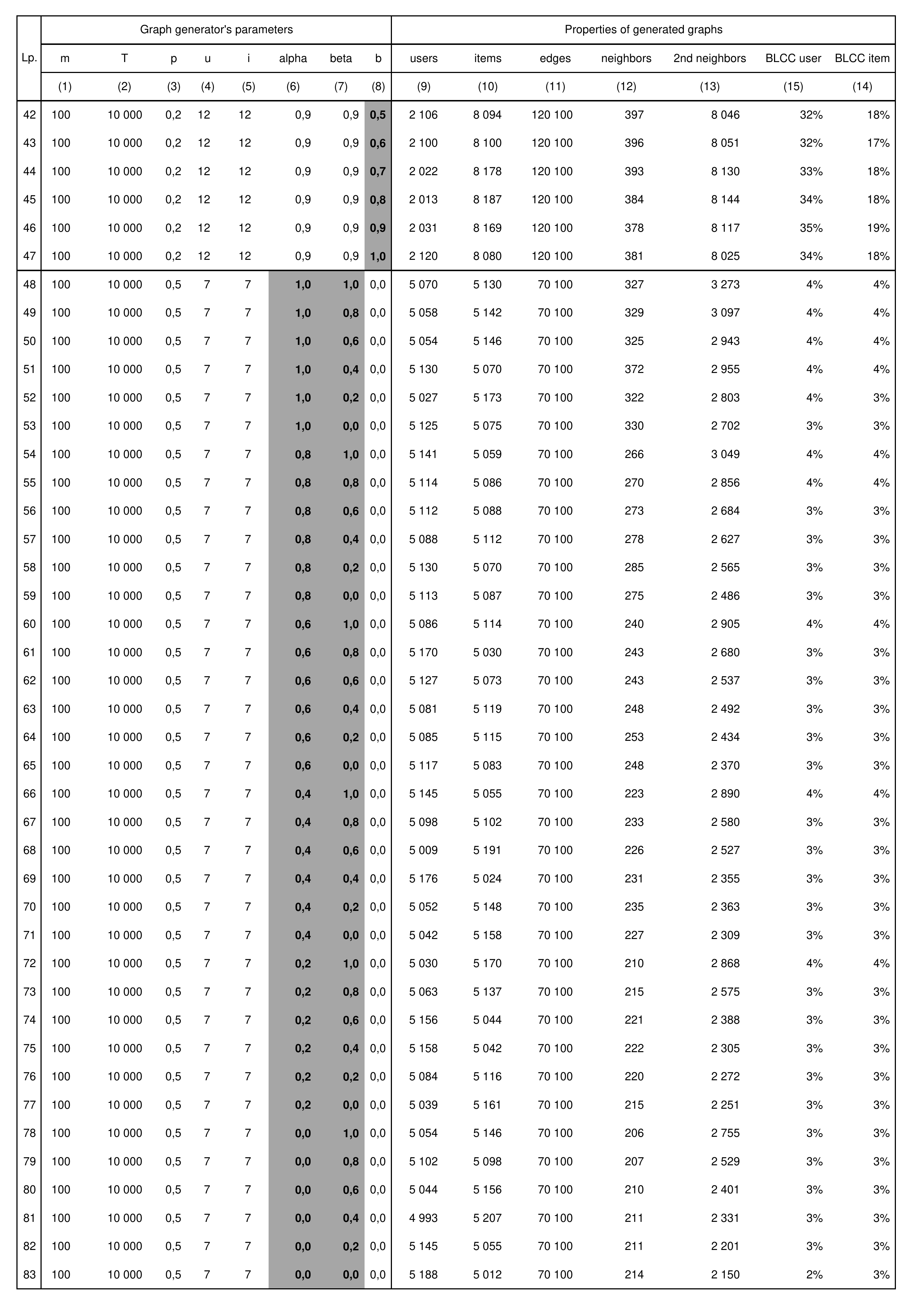}
		\caption{Random graphs generated to test the performance of recommender systems (part 2). Graph generator's parameters and BLCC are defined in Sec 3.1. Neighbors stands for an average number of potentially similar users, $2^{nd}$ neighbors stands for an average number of items of potentially simialar users.}
		\label{fig:tables2}
\end{figure}

We have generated 36 bigraphs with all possible combinations of $\alpha$ and $\beta$ from a set of values $\{0.0, 0.2, 0.4, 0.6, 0.8, 1.0 \}$. All the graphs have $10~000$ nodes and $70~100$ edges. They are numbered from $48$ to $83$ in Fig. \ref{fig:tables2}. We have observed that an average latency grows with either $\alpha$ or $\beta$ for all analyzed algorithms (Fig. \ref{fig:albe2}). We also observed that building time and memory consumption is related to $\alpha$ and $\beta$ for all algorithms except for SVD and KnnItem. The performance of SlopeOne model is drawn in Fig. \ref{fig:albe1}.

\newpage
\subsection{Similarity measure}

In the last two subsections we focus on UserBased algorithm and two parameters that are specific for it. The first parameter is the similarity measure and will be discussed in this subsection. The second parameter is the size of the neighborhood and will be described in next subsection.

The similarity between two vectors $\textbf{x}=(x_1,\dots,x_n)$ and $\textbf{y}=(y_1,\dots,y_n)$ can be defined is several ways. We analyzed all that are available in the Mahout library i.e. \textit{Pearson similarity}, \textit{Euclidean similarity}, \textit{LogLikelihood similarity}, \textit{Spearmann similarity} and \textit{Tanimoto similarity}. The third and the fifth are defined by set operations and can easily by used with \textit{binary ratings}, when we only know if a user expressed a preference for an item or not. \textbf{Pearson similarity} is widely used and is related to the \textit{Cosine similarity}. The only difference is that the former one operates on centered data. \textbf{Euclidean similarity} is negatively proportional to the euclidean distance between two vectors. \textbf{LogLikelihood similarity} is based on calculating four values: number of non-empty dimensions in both vectors, numbers of non-missing dimensions in the first and the second vector, number of all dimensions \cite{loglikelihood}. \textbf{Spearmann similarity} is calculated as Pearson similarity but $x_i$ and $y_j$ are substituted with their relative ranks i.e. the lowest value is $1$, the second lowest is $2$, and so on. \textbf{Tanimoto similarity} for binary data is calculated as a proportion of dimensions that are non-empty in both vectors by total number of non-empty dimensions in the vectors. Detailed definitions of all coefficients can be found in the javadoc API of Mahout \cite{mahout}.

\begin{figure}[htbp]
\centering
	\includegraphics[width=0.5\textwidth]{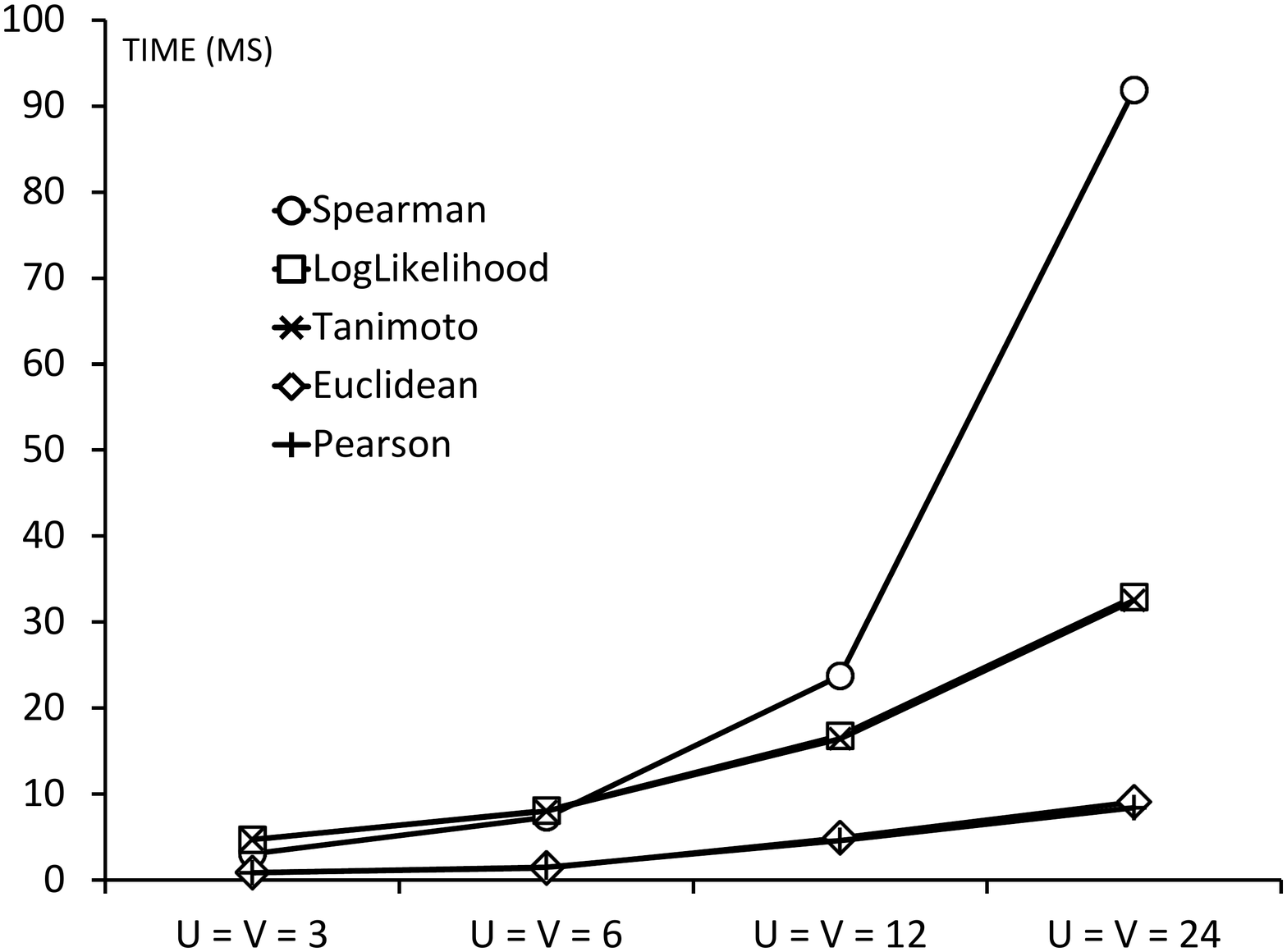}
	
		\caption{Dependence of latency on similarity metrics. Four analyzed graphs have the same number of nodes, they are numbered  $23-26$ in Fig. \ref{fig:tables1}.}
	\label{fig:similarity}
	\end{figure}

We have observed in Fig. \ref{fig:similarity} that the latency of UserBased model depends on the selection of similarity measure. The fastest recommendations are output by Pearson and Euclidean similarities. Tanimoto and Loglikelihood similarities are around three times slower. Spearman similarity is the slowest and does not scale well with the growth of density.

\subsection{Size of the neighborhood}

Size of the neighborhood is a parameter in UserBased recommender, which enables us to tune accuracy of the algorithm. When a UserBased model is requested to deliver recommendations for a given user two steps are performed. In the first step similarity of the user to all other users is calculated and only the most similar users are retained (the number is limited by the neighborhood). In the second step only items of the most similar users are weighted and the recommendation is selected among those items.

\begin{figure}[htbp]
\centering
	\includegraphics[width=0.5\textwidth]{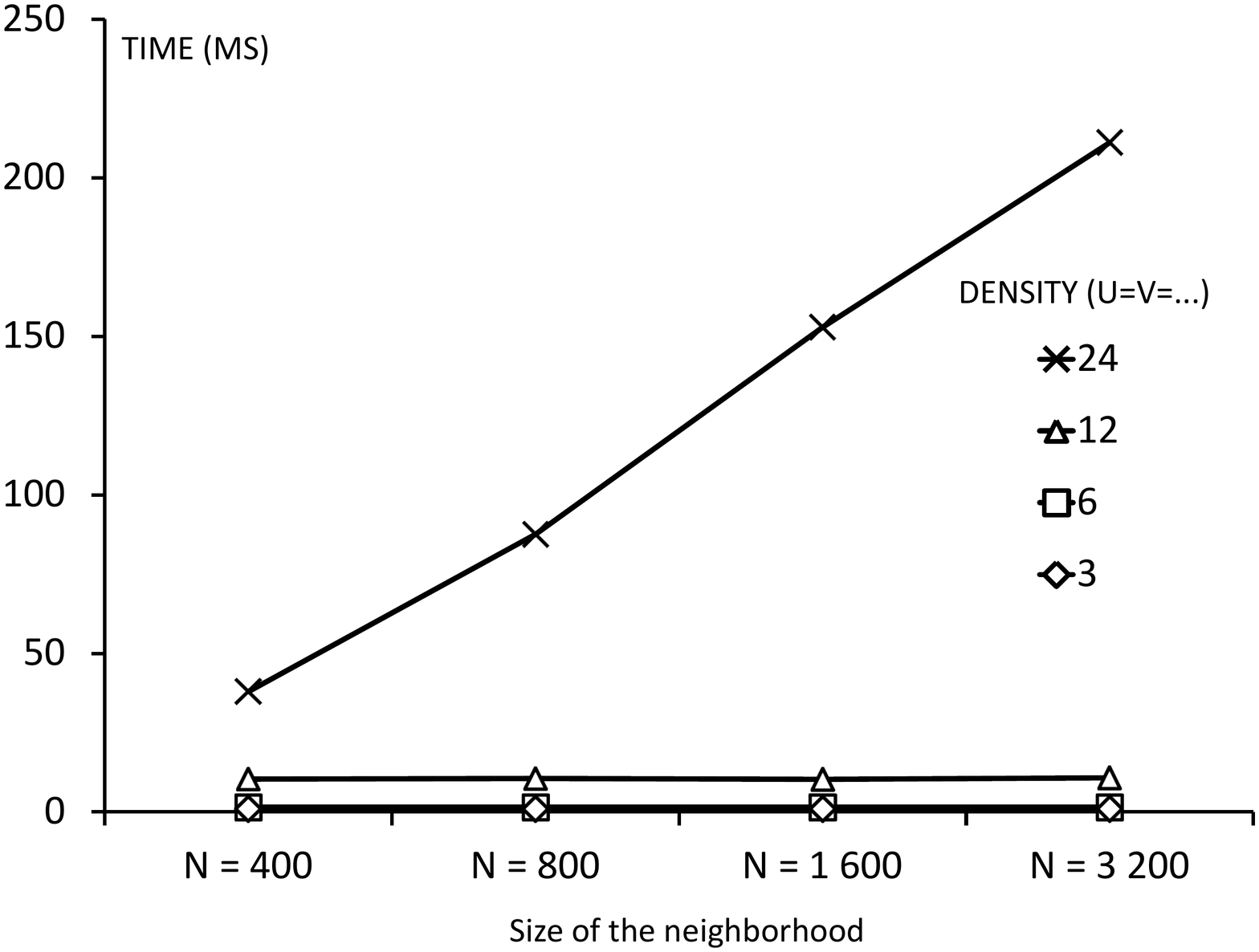}
	\caption{Latency of creating a recommendation in four graphs with varying density. UserBased algorithm run with four different levels of the neighborhood.}
	\label{fig:neighborhood}
\end{figure}

The results in Fig. \ref{fig:neighborhood} indicate that as long as graph's density is below some threshold, the latency does not depend neither on the density nor size of the neighborhood. It can be explained by the fact that the neighborhood parameter is triggered only when the number of similar users is greater than it.  

\section{Conclusion}

In the article we have proposed to use random bipartite graphs to measure the performance of recommender systems. We have showed that recently developed random bigraph generator \cite{chojnacki_CC} can be be used to generate a wide range of artificial datasets with predefined properties. The analytical framework can be used to compare various algorithms, but also to help us understand their complexity and point at potentially non-optimal implementations. We believe that the proposed methodology can be applied in various scenarios and settings. Further analyses need to be performed to understand the most intriguing results. In particular the real relationship between complexity of UserBased and ItemBased algorithms. And the emergence of U-shaped curve describing performance of various recommenders when the proportion of the number of users to the number of items is being changed. In the future we plan to evaluate how the performance of recommender systems is altered when they are implemented in a distributed environment.  

\bibliography{preparation}
\end{document}